%% file: Instruction-Tuned_LLMs_survey.tex
\documentclass[conference]{IEEEtran}
\IEEEoverridecommandlockouts
\usepackage{float}
\usepackage{natbib}
\usepackage{tikz}
\usetikzlibrary{shadows}
\usepackage{float}
\usepackage{forest}
\usetikzlibrary{arrows.meta, shapes.geometric, positioning, fit, backgrounds}
\usepackage{booktabs}
\usepackage{placeins}
\usetikzlibrary{backgrounds}
\usepackage{amsmath,amssymb,amsfonts}
\usepackage{algorithm}
\usepackage{algorithmic}
\usepackage{graphicx}
\usepackage{textcomp}
\usepackage{xcolor}
\usepackage{tcolorbox}
\tcbuselibrary{skins}
\usepackage{enumitem}
\usepackage{longtable}
\usepackage{array}

\tcbset{
  AIbox/.style={
    enhanced,
    colback=gray!3,
    colframe=black!70,
    sharp corners=southwest,
    boxrule=0.4pt,
    left=1em,
    right=1em,
    top=0.8em,
    bottom=0.8em,
    fonttitle=\bfseries,
    title={Comparative Insight: LLaMA-Excitor vs Mosaic-IT}
  }
}

\usepackage{hyperref}
\hypersetup{
    colorlinks=true,
    linkcolor=darkblue,
    citecolor=darkblue,
    urlcolor=darkblue
}
\definecolor{darkblue}{rgb}{0.0, 0.0, 0.55}

\definecolor{foundationblue}{RGB}{70,130,180}
\definecolor{tuningpurple}{RGB}{128,0,128}
\definecolor{scalingorange}{RGB}{255,140,0}
\definecolor{datagreen}{RGB}{34,139,34}
\definecolor{evalred}{RGB}{178,34,34}
\definecolor{ethicsbrown}{RGB}{139,69,19}

\def\BibTeX{{\rm B\kern-.05em{\sc i\kern-.025em b}\kern-.08em
    T\kern-.1667em\lower.7ex\hbox{E}\kern-.125emX}}
\begin{document}

\title{Towards Alignment-Centric Paradigm: A Survey of Instruction Tuning in Large Language Models
\thanks{\hspace*{-\parindent}\rule{3.8cm}{0.4pt} \\ 
$\ast$: Equal contribution. \\
$\dagger$: Corresponding author: Junhao Song (junhao.song23@imperial.ac.uk)}
}

\author{
  \begin{tabular}{c}
    Xudong Han$^{1}$$^{\ast}$, Junjie Yang$^{2}$$^{\ast}$, Tianyang Wang$^{3}$$^{\ast}$, Ziqian Bi$^{4}$, Xinyuan Song$^{5}$, Junfeng Hao$^{6}$, Junhao Song$^{7}$$^{\dagger}$ \\
    \\
    $^{1}$Department of Informatics, University of Sussex, United Kingdom \\
    $^{2}$Pingtan Research Institute, Xiamen University, China \\
    $^{3}$Department of Computer Science, University of Liverpool, United Kingdom \\
    $^{4}$Department of Computer Science, Purdue University, United States \\
    $^{5}$Department of Computer Science, Emory University, United States \\
    $^{6}$AI Agent Lab, Vokram Group, United Kingdom \\
    $^{7}$Department of Computing, Imperial College London, United Kingdom \\
    \texttt{xh218@sussex.ac.uk, youngboy@xmu.edu.cn, tianyangwang0305@gmail.com,} \\
    \texttt{bi32@purdue.edu, xsong30@emory.edu, ai-agent-lab@vokram.com,} \\
    \texttt{junhao.song23@imperial.ac.uk$^{\dagger}$}
  \end{tabular}
}

\maketitle
\begin{abstract}
Instruction tuning is a pivotal technique for aligning large language models (LLMs) with human intentions, safety constraints, and domain-specific requirements. This survey provides a comprehensive overview of the full pipeline, encompassing (i) data collection methodologies, (ii) full-parameter and parameter-efficient fine-tuning strategies, and (iii) evaluation protocols. We categorized data construction into three major paradigms: expert annotation, distillation from larger models, and self-improvement mechanisms, each offering distinct trade-offs between quality, scalability, and resource cost. Fine-tuning techniques range from conventional supervised training to lightweight approaches, such as low-rank adaptation (LoRA) and prefix tuning, with a focus on computational efficiency and model reusability. We further examine the challenges of evaluating faithfulness, utility, and safety across multilingual and multimodal scenarios, highlighting the emergence of domain-specific benchmarks in healthcare, legal, and financial applications. Finally, we discuss promising directions for automated data generation, adaptive optimization, and robust evaluation frameworks, arguing that a closer integration of data, algorithms, and human feedback is essential for advancing instruction-tuned LLMs. This survey aims to serve as a practical reference for researchers and practitioners seeking to design LLMs that are both effective and reliably aligned with human intentions.
\end{abstract}


\section{Foundations and Instruction Tuning of Large Language Models}

Large Language Models (LLMs) have dramatically transformed natural language processing, bringing new capabilities for text generation, comprehension, and analytical tasks. The rapid evolution of LLM research has resulted in various applications, including chatbots, virtual assistants, content generation, and translation systems \citep{2307.06435}. However, despite these impressive performances, LLMs still have constraints that require careful understanding of their underlying principles, advantages, and limitations to properly utilize their potential \citep{2412.04503}.

A thorough understanding of LLMs is essential for gaining deeper insights. Recent studies have offered valuable perspectives on their historical evolution, development trajectory, and core principles \citep{2402.06853}. The significance of fine-tuning and training approaches for LLMs has been highlighted, with empirical research showing that supervised fine-tuning works well for smaller-scale LLMs \citep{2412.13337}. Additionally, the considerable computational resources and expertise required for LLM development create obstacles for practitioners, emphasizing the necessity of efficient training setups and methodologies \citep{2308.10252}.

\begin{figure*}[t]
    \centering
    \includegraphics[width=0.8\linewidth]{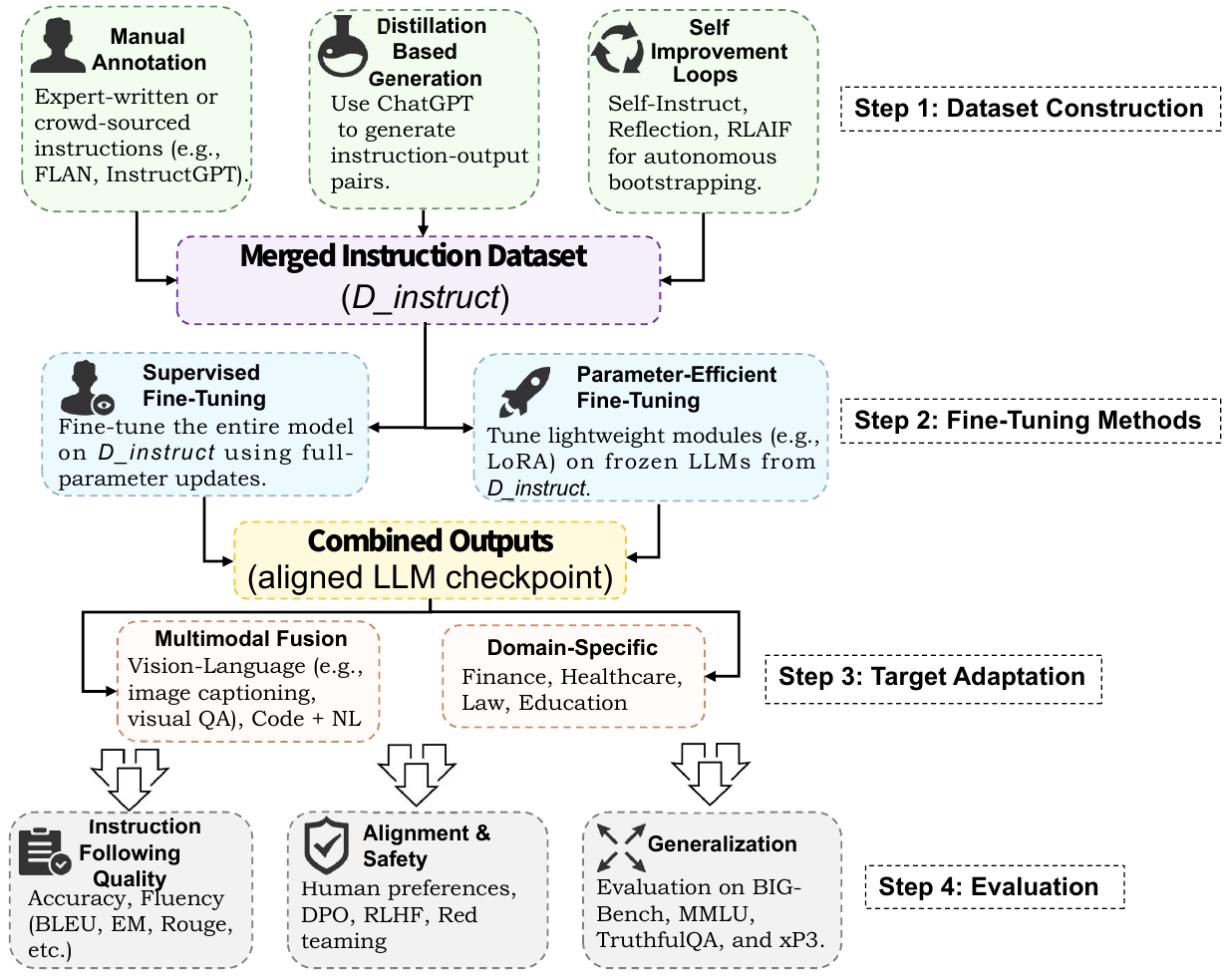}
    \caption{
    Overview of the instruction-tuning pipeline for large language models (LLMs). 
    The process begin with dataset construction from three main sources: manual annotation, distillation-based generation, and self-improvement loops. These are merged into a unified instruction dataset \( D_{\text{instruct}} \), which is then utilized in either supervised fine-tuning or parameter-efficient fine-tuning (e.g., LoRA \citep{hu2021lora}, Prefix \citep{li2021prefix}). 
    The resulting model is adapted for multi-modal or domain-specific tasks, followed by evaluation across three dimensions: instruction-following quality, alignment and safety, and generalization. 
    }
    \label{fig:instruction_tuning_pipeline}
\end{figure*}

The present state of LLM research is marked by rapid progress, with novel architectures, training techniques, and use cases constantly appearing \citep{2401.13601}. Multi-modal LLMs that combine text with other modalities, such as images or audio, have demonstrated strong potential in areas such as visual question answering and multi-modal dialogue systems \citep{2310.00492}. In addition, the creation of specialized tools and frameworks, such as LLMBox, has made it easier to deploy and integrate LLMs into practical applications \citep{2407.05563}.

However, multiple limitations and challenges remain, such as the requirement for massive training datasets, substantial computational power, and deep learning expertise \citep{2408.04643}. Moreover, concerns about the risks and unexpected effects of broad LLM deployment have been raised, stressing the importance of responsible AI development practices \citep{2305.09246}. Research on instruction tuning for LLMs also emphasizes the need to understand the behavioral changes in LLMs following instruction tuning \citep{2310.00492}.

Instruction tuning addresses the key shortcomings of pretrained LLMs, which often generate convincing but incorrect answers to complicated or domain-specific questions. By exposing models to carefully selected datasets pairing instructions with expected outputs, instruction tuning enhances task performance, reasoning capabilities, and human intent alignment. \citep{2310.00492} highlights it's crucial importance in sensitive areas like healthcare and finance where accuracy are essential. This procedure involves building high-quality datasets via manual annotation, distillation, or self-improvement approaches, each of which has different trade-offs regarding quality versus scalability.

In summary, the rise of large language models has fundamentally changed natural language processing, enabling text understanding, generation, and reasoning abilities that were previously impossible. These models have accelerated progress in many applications, from conversational systems to scientific research, demonstrating their transformative impact. However, widespread LLM deployment requires a thorough understanding of their basic principles, design choices, and behavioral constraints. To maximize their benefits, it is important that researchers and engineers work systematically to tackle the technical, computational, and ethical issues that come with their usage \citep{2304.13712}. As this field continues to evolve rapidly, future advances will depend not only on architectural improvements but also on the responsible implementation of instruction tuning and alignment methods. These efforts should be directed by thoughtful evaluation frameworks and social considerations, ensuring that LLMs are built with transparency, safety, and broad applicability in mind \citep{2402.06853}.


\section{Instruction Tuning Pipeline and Formalization}

The instruction-tuning pipeline for large language models represents a structured way to improve model abilities by learning from human demonstrations and preferences. As shown in \textbf{Figure~\ref{fig:instruction_tuning_pipeline}}, this complete framework includes four main stages: dataset creation, fine-tuning approaches, target adaptation, and evaluation measures.

\noindent\textbf{Dataset Construction (Step 1):} The basis of instruction tuning lays in building quality instruction datasets using three main methods. Manual annotation involves expert-created or crowdsourced instructions, as shown in datasets such as FLAN \citep{wei2021finetuned} and InstructGPT \citep{ouyang2022training}. These datasets usually contain instruction-response pairs $(x_i, y_i)$ where $x_i$ represents the natural language instruction and $y_i$ is the expected response. The quality of manually annotated data is formalized as follows:
\begin{equation}
\mathcal{Q}_{\text{manual}} = \frac{1}{N} \sum_{i=1}^{N} \mathbb{I}[\text{Human-Judge}(x_i, y_i) \geq \tau],
\end{equation}
where $\mathbb{I}[\cdot]$ is the indicator function, $\tau$ represents the quality threshold, and $N$ is the total number of instruction-response pairs.

Distillation-based generation uses powerful models, such as ChatGPT, to automatically create instruction-output pairs. This method follows the knowledge distillation approach, where a teacher model $M_{\text{teacher}}$ generates responses for a given instruction set $\mathcal{X}$:
\begin{equation}
\mathcal{D}_{\text{distill}} = \{(x_i, M_{\text{teacher}}(x_i)) \mid x_i \in \mathcal{X}\},
\end{equation}
The distillation process aims to minimize the distribution divergence between the teacher and student models:
\begin{equation}
\mathcal{L}_{\text{distill}} = \mathbb{E}_{x \sim \mathcal{X}} \left[ \text{KL}(P_{\text{teacher}}(\cdot|x) \| P_{\text{student}}(\cdot|x)) \right].
\end{equation}

Self-improvement loops form the third component of dataset building, using techniques such as Self-Instruct \citep{wang2022self}, Constitutional AI \citep{bai2022training}, and Reinforcement Learning from AI Feedback (RLAIF) \citep{bai2022constitutional}. These approaches allow autonomous bootstrapping via iterative improvements.
\begin{equation}
\mathcal{D}_{t+1} = \mathcal{D}_t \cup \{(x_i, \text{Improve}(M_t(x_i))) \mid x_i \in \mathcal{S}_t\},
\end{equation}
where $\mathcal{S}_t$ represents the seed instruction set at iteration $t$, and $\text{Improve}(\cdot)$ is the self-improvement function.

The self-improvement process involves combining different data sources into one unified instruction dataset, which is mathematically expressed as:
\begin{equation}
D_{\text{instruct}}
= \mathcal{D}_{\text{manual}} 
\cup \mathcal{D}_{\text{distill}} 
\cup \mathcal{D}_{\text{self-improve}} .
\end{equation}

\noindent\textbf{Fine-Tuning Methods (Step 2).} 
The instruction dataset serves as the input for two primary fine‑tuning paradigms.  
\emph{Supervised Fine‑Tuning (SFT)} performs full‑parameter updates across the entire model architecture.  
To avoid length bias, the loss is normalized by the total number of target tokens rather than the number of samples:
\small
\begin{equation}
\mathcal{L}_{\text{SFT}}
= -\frac{1}
       {\displaystyle\sum_{(x,y)\,\in\,D_{\text{instruct}}}|y|}
     \;\sum_{(x,y)\,\in\,D_{\text{instruct}}}
     \;\sum_{t=1}^{|y|}
     \log P_{\theta}\!\bigl(y_t \mid x,\,y_{<t}\bigr),
\end{equation}
where $\theta$ denotes the model parameters, $y_t$ is the $t$‑th token in the response, and $y_{<t}$ is the preceding target tokens.

Parameter-Efficient Fine-Tuning (PEFT) techniques, including Low-Rank Adaptation (LoRA) \citep{hu2021lora}, Prefix Tuning \citep{li2021prefix}, and adapters \citep{houlsby2019parameter}, adjust only small task-specific parts while keeping pre-trained weights frozen. For LoRA, the weight update is decomposed as:
\begin{equation}
W' = W_0 + \Delta W = W_0 + BA,
\end{equation}
where $W_0 \in \mathbb{R}^{d \times k}$ represents the frozen pre-trained weights, $B \in \mathbb{R}^{d \times r}$, $A \in \mathbb{R}^{r \times k}$, and $r \ll \min(d,k)$ are the low-rank bottleneck dimensions.

The computational efficiency of the LoRA method can be measured as follows:
\begin{equation}
\text{Efficiency Ratio} = \frac{|\theta_{\text{trainable}}|}{|\theta_{\text{total}}|} = \frac{r(d + k)}{dk} = r\left(\frac{1}{d} + \frac{1}{k}\right),
\end{equation}
where \( r \) is the rank of the low-rank matrices, \( d \) is the input feature dimension, and \( k \) is the output feature dimension. The trainable parameters are those in matrices \( B \in \mathbb{R}^{d \times r} \) and \( A \in \mathbb{R}^{r \times k} \), which are used to approximate the original weight matrix \( W \in \mathbb{R}^{d \times k} \) via the product \( BA \).

\noindent\textbf{Target Adaptation (Step 3):} The fine-tuned models undergo specialized adaptation for specific applications. Multi-modal fusion integrates vision-language capabilities for tasks such as image captioning, visual question answering, and code generation using natural language. The multi-modal objective combines text and visual representations as:
\begin{equation}
\mathcal{L}_{\text{multimodal}} = \mathcal{L}_{\text{text}} + \lambda_v \mathcal{L}_{\text{vision}} + \lambda_f \mathcal{L}_{\text{fusion}},
\end{equation}
where $\lambda_v$ and $\lambda_f$ are the weighting coefficients for the vision and fusion losses, respectively.

Domain-specific adaptation customizes models for specialized fields, such as finance, healthcare, law, and education. The domain adaptation loss includes domain-specific terms and reasoning patterns as follows:
\begin{equation}
\mathcal{L}_{\text{domain}} = \mathcal{L}_{\text{general}} + \alpha \mathcal{L}_{\text{domain-specific}} + \beta \mathcal{L}_{\text{domain-consistency}}.
\end{equation}
where $\alpha$ and $\beta$ control the balance between general capabilities and the domain knowledge.

\noindent\textbf{Evaluation Framework (Step 4):} The instruction-tuned models are evaluated across three crucial dimensions. Instruction-Following Quality measures the model's ability to comprehend and properly execute the provided instructions. Common metrics include BLEU \citep{papineni2002bleu}, ROUGE \citep{lin2004rouge}, and BERTScore \citep{zhang2019bertscore}:
\begin{equation}
\text{IF-Score} = \frac{1}{N} \sum_{i=1}^{N} \text{Similarity}(\text{Response}_i, \text{Reference}_i).
\end{equation}

Alignment and Safety evaluation cover human preference alignment utilizing techniques such as Direct Preference Optimization (DPO) \citep{rafailov2023direct}, Reinforcement Learning from Human Feedback (RLHF) \citep{christiano2017deep}, and red teaming protocols. The alignment objective is expressed as:
\begin{equation}
\begin{split}
\mathcal{L}_{\text{alignment}} = -\mathbb{E}_{(x,y_w,y_l) \sim \mathcal{D}} \bigg[ 
\log \sigma\Big( \beta \log \frac{\pi_{\theta}(y_w|x)}{\pi_{\text{ref}}(y_w|x)} \\
\quad - \beta \log \frac{\pi_{\theta}(y_l|x)}{\pi_{\text{ref}}(y_l|x)} \Big) \bigg]
\end{split},
\end{equation}
where $y_w$ and $y_l$ represent preferred and dispreferred responses, respectively; $\pi_{\text{ref}}$ is the reference model; and $\beta$ is the temperature parameter.

Generalization assessment evaluates model performance across diverse benchmarks including BIG-Bench \citep{srivastava2022beyond}, MMLU \citep{hendrycks2020measuring}, TruthfulQA \citep{lin2021truthfulqa}, and multilingual tests like XCOPA \citep{ponti2020xcopa} and MT-Bench \citep{zheng2023judging}. The generalization capability is quantified by cross-domain performance consistency as follows:
\begin{equation}
\text{Generalization Score} = \frac{1}{|\mathcal{D}|} \sum_{d \in \mathcal{D}} \text{Performance}(M, \mathcal{T}_d),
\end{equation}
where $\mathcal{D}$ represents the evaluation domain set and $\mathcal{T}_d$ is the task set for domain $d$.

This comprehensive pipeline ensures the systematic development of instruction-following capabilities while maintaining alignment with human values and demonstrating robust generalization across various applications. The combination of diverse data sources, flexible fine-tuning techniques, targeted adaptation methods, and comprehensive evaluation protocols establishes a structured framework for enhancing the abilities of large language models through instruction tuning.

\begin{figure*}[t]
    \centering
    \includegraphics[width=\linewidth]{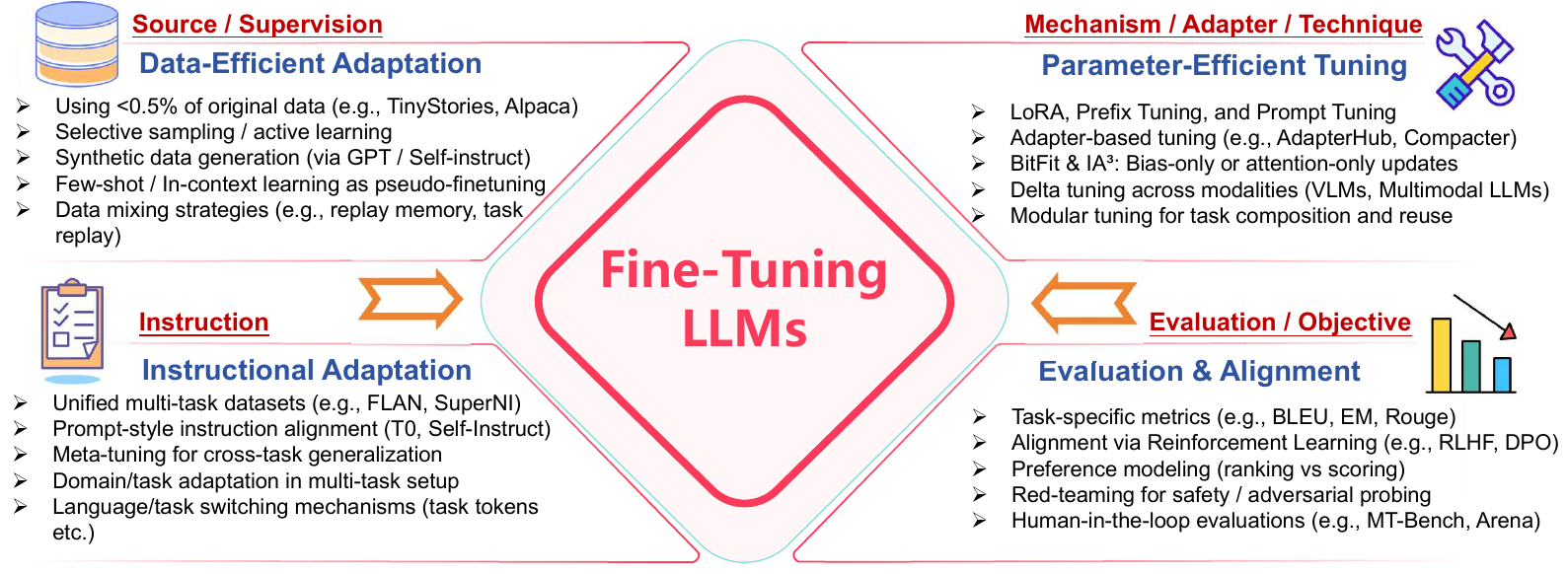}
    \caption{A structured taxonomy of strategies for fine-tuning large language models (LLMs), organized into four primary directions: data-efficient adaptation, parameter-efficient tuning, instructional adaptation, and evaluation/alignment. Each block highlight representative approaches and techniques.}
    \label{fig:llm_finetuning_taxonomy}
\end{figure*}

Large language models (LLMs) are built on diverse architectural designs, training methodologies, and numerous practical applications. Among these approaches, instruction tuning has emerged as a crucial technique with various benefits and limitations. \citep{2402.05119} demonstrated that instruction tuning may not necessarily enhance the fundamental knowledge or capabilities of LLMs and might actually lead to knowledge degradation. Conversely, \citep{2310.00492} investigated the internal transformations resulting from instruction tuning and discovered enhancements in instruction recognition, semantic alignment, and reorienting pretrained knowledge toward task-specific goals.

The efficiency of the model design has received considerable attention. \citep{2409.01990} presented comprehensive analysis of efficient training and inference techniques, proposing model and system co-design for reducing computational expenses and improving accessibility. Similarly, \citep{2307.06435} offered a broader survey of LLM research, covering recent architectural innovations, training improvements, and multi-modal modeling developments.

Model evaluation remains a critical aspect of LLM development. To address this challenge, \citep{2406.00936} proposed a two-stage evaluation framework that assesses both the fundamental capabilities and agent-level implementations of LLMs. This emphasizes the growing necessity to elucidate model abilities and establish suitable task boundaries. Regarding security and behavioral alignment, \citep{2410.04524} explored how instruction tuning could potentially introduce vulnerabilities, highlighting the requirement for secure and interpretable fine-tuning methods.

Rigorous experimental designs have been utilized to comprehend LLM behavior from a methodological standpoint. For instance, \citep{2402.05119} performed controlled experiments to investigate the limitations of instruction tuning, whereas \citep{2310.00492} created local and global interpretation methods for analyzing attention mechanisms and internal representations. Furthermore, \citep{2409.01990} stressed the importance of collaborative optimization across software and hardware to enhance training scalability.

Multiple unresolved challenges continue to influence the research landscape. The limitations of instruction tuning, especially its restricted efficacy in knowledge augmentation, persist as open problems \citep{2402.05119}. Efficient model deployment requires overcoming computational obstacles, and evaluation frameworks need to evolve to measure subtle capabilities and responsibilities within dynamic tasks. Comprehending post-tuning behavioral changes also represents a vital area for subsequent investigation, particularly for high-stakes deployments.

Collectively, recent research offers multi-dimensional insights into LLM development, spanning architecture, optimization, and task alignment. Instruction tuning, model co-design, and evaluation approaches serve as the primary drivers of contemporary practice. These contributions emphasize the necessity for ongoing research on robustness, scalability, and cross-domain synthesis.

\section{Fine-Tuning Methodologies and Scalable Pipelines for LLMs}

Fine-tuning large language models (LLMs) for specific tasks has emerged as a central topic in recent research, offering opportunities for enhanced accuracy, adaptability, and computational efficiency in various application domains. This process involves modifying pretrained LLM parameters to align more closely with target task requirements, often yielding superior results compared to general-purpose models \citep{2305.09246}. Effective prompt engineering strategies (see \textbf{Table~\ref{tab:prompt_engineering}}) can augment these fine-tuning approaches to further boost model performance.

\sloppy
In a typical fine-tuning workflow, let the pretrained model parameters be $\theta_{0}$, and let the target task dataset be $\{(x_{i}, y_{i})\}_{i=1}^{N}$. We update the parameters by minimizing the token-level cross-entropy loss \citep{radford2018improving, brown2020language, raffel2020exploring}:
\begin{equation}
  \mathcal{L}_{\mathrm{FT}}(\theta)
  \;=\;
  \frac{1}{\displaystyle\sum_{i=1}^{N} T_{i}}
  \sum_{i=1}^{N} \sum_{t=1}^{T_{i}}
    -\log p_{\theta}\!\bigl(y_{i}^{(t)} \mid y_{i}^{(<t)},\,x_{i}\bigr)\,,
  \label{eq:ft_obj}
\end{equation}
where each $y_{i}$ is a sequence of target tokens of length $T_{i}$, and
$p_{\theta}(\cdot)$ denotes the predicted probability distribution of the LLM over the vocabulary. This formulation follows the standard practice in autoregressive fine-tuning \citep{brown2020language}, where the likelihood of the ground-truth token at each time step is maximized given all previous tokens.   
\fussy

One of the primary challenges in this domain is the reliance on extensive, high-quality training data. Nevertheless, recent findings suggest that task-specific fine-tuning can achieve high effectiveness, even with significantly smaller data volumes. For instance, models have been successfully adapted using less than 0.5\% of the original dataset while preserving or enhancing task performance \citep{2305.09246}. These results indicate that with thoughtfully selected or curated data, fine-tuning becomes feasible in low-resource or specialized contexts, thus expanding its applicability across diverse fields. When dealing with such limited datasets, the structured prompting methods described in \textbf{Table~\ref{tab:prompt_engineering}} are especially useful for optimizing model performance.

Suppose that we extract only $M$ samples from the original corpus $\mathcal{D}_{\text{orig}}$ with $|\mathcal{D}_{\text{orig}}| = N_{\text{orig}}$ for fine-tuning, where $M \ll N_{\text{orig}}$. The sampling ratio can be formalized as:
\begin{align}
\mathcal{D}_{\text{sample}} &= \text{Sample}(\mathcal{D}_{\text{orig}}, M), \label{eq:data_sampling_a}\\
M &= \alpha \cdot N_{\text{orig}}, \quad \alpha \ll 1.
\label{eq:data_sampling_b}
\end{align}

To augment the limited sample set $\mathcal{D}_{\text{sample}} = \{x_1, x_2, \ldots, x_M\}$, synthetic data can be generated using various augmentation strategies.
\begin{equation}
\mathcal{D}_{\text{aug}} = \{g_{\text{aug}}(x_i, \theta) \mid x_i \in \mathcal{D}_{\text{sample}}\},
\label{eq:synthetic_data}
\end{equation}
where $g_{\text{aug}}(\cdot, \theta)$ represents the augmentation function, which may include token-level augmentation (random masking, synonym replacement), semantic augmentation (paraphrasing, back-translation), or generative augmentation using large language models.

The final training dataset combines the original samples with the augmented data:
\begin{align}
\mathcal{D}_{\text{train}} &= \mathcal{D}_{\text{sample}} \cup \mathcal{D}_{\text{aug}}, \label{eq:final_dataset_a}\\
|\mathcal{D}_{\text{train}}| &= M + |\mathcal{D}_{\text{aug}}|.
\label{eq:final_dataset_b}
\end{align}

As illustrated in \textbf{Figure~\ref{fig:llm_finetuning_taxonomy}}, contemporary fine-tuning strategies for large language models (LLMs) can be systematically organized into four principal categories, each addressing a specific adaptation dimension. The first category, \textit{Data-Efficient Adaptation}, emphasizes maximizing performance using minimal training data via methods such as sub-sampling under 0.5\% of the original data (e.g., TinyStories \citep{Eldan2023TinyStories}, Alpaca), active data selection, synthetic data generation through GPT or Self-Instruct, few-shot or in-context learning as implicit fine-tuning, and data mixing approaches including task replay and memory-based scheduling. The second category, \textit{Parameter-Efficient Tuning}, focuses on decreasing the number of trainable parameters while preserving task adaptability, including methods such as LoRA \citep{hu2021lora}, prefix \citep{li2021prefix}, and prompt tuning \citep{lester2021power}, adapter-based architectures (e.g., AdapterHub \citep{pfeiffer2020adapterhub}, Compacter \citep{karimi2021compacter}), and lightweight bias- or attention-only tuning (e.g., BitFit \citep{zaken2021bitfit}, IA$^3$ \citep{liu2022few}), frequently allowing modular reuse throughout tasks or modalities. The third category, \textit{Instructional Adaptation}, address the challenge of generalization across varied tasks through utilizing unified multi-task datasets (e.g., FLAN \citep{wei2021finetuned}, SuperNI \citep{wang2022super}), prompt-style instruction tuning frameworks (e.g., T0 \citep{sanh2021multitask}, Self-Instruct \citep{wang2022self}), meta-tuning for enhanced transferability \citep{yin2020meta}, domain-aware task conditioning \citep{bai2024efficient}, and mechanisms like task tokens for switching among objectives. Finally, \textit{Evaluation and Alignment} encompass both quantitative and qualitative assessments of LLM behavior, spanning from task-specific metrics (e.g., BLEU \citep{papineni2002bleu}, EM \citep{rajpurkar2016squad}, Rouge \citep{lin2004rouge}) to alignment through reinforcement learning approaches (e.g., RLHF \citep{ouyang2022training}, DPO \citep{rafailov2023direct}), preference modeling (ranking vs scoring) \citep{wirth2017survey}, safety probing using red-teaming \citep{perez2022red}, and human-in-the-loop evaluation platforms including MT-Bench \citep{zheng2023judging} and Arena. Combined, these four directions offer a comprehensive taxonomy of strategies for enhancing the efficiency, flexibility, and safety of adapting foundation models for downstream tasks.

Efficient fine-tuning techniques have become a primary focus for adapting large language models (LLMs) to particular tasks with reduced computational expense. Recent methods have presented the synchronized tuning of prompts and prefixes, along with feature interaction modules that preserve pre-trained knowledge while improving adaptation abilities \citep{2402.01643, 2404.00913}. These approaches usually involve modifying or supplementing model components to enhance performance and stability throughout task-specific fine-tuning. For example, indirect feature interactions have been demonstrated to facilitate more effective knowledge alignment among pretrained representations and novel tasks \citep{2404.00913}.

Prefix-tuning \citep{li2021prefix} is a parameter-efficient adaptation technique that introduces trainable continuous vectors, referred to as \textit{prefixes}, which are prepended to the input sequence at each Transformer layer. Unlike the direct modification of attention matrices, prefix-tuning optimizes task-specific prefix parameters $\mathbf{P}_{\text{task}}$ that are concatenated with the input embeddings.

Formally, for the input sequence $\mathbf{X} = [\mathbf{x}_1, \mathbf{x}_2, \ldots, \mathbf{x}_n]$, the prefixed input at layer $l$ becomes:
\begin{equation}
\mathbf{X}_l' = [\mathbf{P}_l; \mathbf{X}_l],
\label{eq:prefix_input}
\end{equation}
where $\mathbf{P}_l \in \mathbb{R}^{|P| \times d}$ represents the trainable prefix vectors for layer $l$, $|P|$ is the prefix length, and $d$ is the hidden dimension. The attention mechanism then operates over this extended sequence as follows:
\begin{align}
\mathbf{Q}', \mathbf{K}', \mathbf{V}' &= f_{\text{proj}}(\mathbf{X}_l'), \label{eq:prefix_proj_a}\\
\text{Attn} &= \text{Softmax}\left(\frac{\mathbf{Q}' {\mathbf{K}'}^T}{\sqrt{d}}\right) \mathbf{V}',
\label{eq:prefix_proj_b}
\end{align}
where $f_{\text{proj}}$ denotes the linear projection layers for queries, keys, and values. During training, only the prefix parameters $\{\mathbf{P}_l\}_{l=1}^L$ are updated, while the pre-trained Transformer parameters remain frozen, achieving parameter efficiency with typically $<1\%$ of the original model parameters.

Multi-task instruction tuning represents another promising methodology that enables single models to manage multiple related tasks via shared supervision. This strategy enhances generalization and robustness by exposing models to varied objectives, frequently yielding improved performance in task-specific implementations \citep{2305.13225}. By learning shared representations among tasks, the models can transfer knowledge more efficiently and minimize overfitting within low-resource settings.

The evaluation of fine-tuned LLMs presents distinct challenges, especially in determining suitable metrics that reflect both task accuracy and instruction compliance. Research has highlighted the significance of aligning evaluation criteria with practical use cases, particularly for industrial implementations \citep{2310.12345}. The selection of metrics substantially impacts the assessment of the fine-tuning method’s effectiveness, emphasizing the necessity for standardized, domain-aware evaluation protocols.

Current debates in this field reflect divergent viewpoints regarding fine-tuning strategies. While certain efforts have focused on direct tuning using extensive datasets, others have investigated indirect approaches that reduce data demands and emphasize parameter efficiency \citep{2305.09246}. Moreover, the trade-offs between single- and multi-task tuning continue to shape methodological choices, with various approaches providing benefits based on domain and resource limitations \citep{2402.01643}.

As summarized in \textbf{Table~\ref{tab:dataset_creation_comparison}}, Manual Construction delivers the highest semantic fidelity owing to expert-authored instructions; however, its heavy reliance on human labor limits scalability. Distillation-based methods enable the rapid, moderate-cost generation of large-scale instruction data, although the quality remains bounded by the teacher model’s capability and prompt engineering. Self-improvement approaches leverage iterative model self-critique to enhance scalability and achieve competitive quality in subsequent iterations.  

\begin{table*}[ht!]
\centering
\caption{
Comparison of instruction-tuning dataset-creation approaches. 
\newline
\small Symbols: $\checkmark$ = strength, $\times$ = limitation, $\sim$ = moderate, $\triangle$ = variable; see footnotes for methodology specifics and examples.
}
\label{tab:dataset_creation_comparison}
\begin{tabular}{p{4.6cm}p{5.2cm}p{5.2cm}}
\toprule
\textbf{Approach} & \textbf{Likely Quality of Instructions} & \textbf{Likely Scalability of Dataset Generation} \\
\midrule
Manual Construction$^{\dagger}$ & 
High $\checkmark$ \newline
{\footnotesize Human-authored instructions are typically accurate, diverse and contextually grounded. These reflect real-world intentions, user expectations, and domain expertise. Such instructions are well-structured, interpretable, and have high semantic fidelity. Safety considerations and the avoidance of offensive or harmful content are often included. Used in datasets like InstructGPT \citep{ouyang2022training}, FLAN \citep{wei2021finetuned}, and T5's C4 prompt tuning \citep{raffel2020exploring}.} &
Low $\times$ \newline
{\footnotesize Creation is labor-intensive and requires expert writers or crowd workers with detailed guidelines. Quality control (e.g., filtering toxic or low-quality examples) further increases the costs. Scaling beyond tens of thousands of examples is rarely feasible. Limited adaptability across new domains due to human bottlenecks.} \\

Distillation-based$^{\ast}$ & 
Medium $\sim$ \newline
{\footnotesize Instructions synthesized by prompting large teacher models (e.g., GPT-4 \citep{openai2023gpt4}) are often diverse and grammatically fluent. However, quality is closely linked to the teacher’s model and prompt design. However, issues such as hallucinations, verbosity, and misalignment may arise. Semantic fidelity is moderate and varies according to the filtering strategy. Human post-editing is usually needed to ensure correctness and safety.} & 
Medium $\sim$ \newline
{\footnotesize Can rapidly generate large volumes using LLM APIs with prompt chaining and task variation techniques. The efficiency depends on the cost and availability of the teacher models. Scalable with a computational budget, risks amplify model biases and generate inconsistent outputs. Some domain coverage limitations depending on the teacher’s knowledge.} \\

Self-improvement / Bootstrapping$^{\ddagger}$ &
Variable $\triangle$ \newline
{\footnotesize Involves autonomous generation and refinement of instruction–response pairs by the model itself, often with self-critique, feedback, or ranking mechanisms (e.g.,  reflection \citep{openai2023reflection}, RLAIF \citep{lee2023rlaif}, LLM-as-a-judge \citep{zhang2023llmjudge}). The quality varied significantly during the early iterations. Improvements gradually emerged with more refinement loops being added. Instructions may be creative or domain-specific, but also noisy or inconsistent without supervision.} & 
High $\checkmark$ \newline
{\footnotesize Facilitates continual, large-scale instruction generation without reliance on external models or humans. It is suited for long-term domain-adaptive training. It enables learning new tasks or styles through recursive feedback. Although the initial samples may be unstable, later iterations can achieve high quality. Can be paired with minimal human filtering or reward models to boost safety.} \\
\bottomrule
\end{tabular}
\vspace{0.8em}
\begin{minipage}{\textwidth}
\footnotesize
\textbf{Notes:} \\
$^{\dagger}$ \textit{Manual Construction} refers to human-written instruction–response datasets produced by experts or crowd workers, typically using annotation pipelines and quality assurance steps. Representative datasets include InstructGPT’s preference-tuned instructions \citep{ouyang2022training}, FLAN’s task collection \citep{wei2021finetuned}, and Super-Natural Instructions \citep{wang2022supernatural}. \\
$^{\ast}$ \textit{Distillation-based} methods involve prompting strong proprietary models (e.g., GPT-3.5/4) to generate synthetic instruction–response pairs, followed by deduplication, ranking, and heuristic filtering. Notable examples include Stanford Alpaca \citep{taori2023stanfordalpaca}, Evol-Instruct \citep{xu2023wizardlm}, and Vicuna \citep{chiang2023vicuna}. Techniques such as task mutation and prompt chaining improve diversity. \\
$^{\ddagger}$ \textit{Self-improvement/Bootstrapping} involves self-play, recursive self-critique, or environmental interaction to generate, evaluate, and refine instructions. Examples include reflection (feedback-aware planning) \citep{openai2023reflection}, RLAIF (Reinforcement Learning with AI Feedback) \citep{lee2023rlaif}, AutoGPT-style agents \citep{yang2023auto}, and LLM-as-a-judge systems \citep{zhang2023llmjudge}. These pipelines emphasize autonomy and long-horizon learning potential.
\end{minipage}
\end{table*}

Despite these advances, several key limitations persist. Access to high-quality domain-specific data remains a significant bottleneck, especially for under-resourced languages or specialized use cases \citep{2305.09246}. Additionally, scalability and efficiency continue to be critical concerns because fine-tuning large models often requires considerable computational resources and infrastructure \citep{2402.01643}.

Recent progress in fine-tuning methodologies, multi-task training paradigms, and evaluation approaches has substantially advanced this field. While challenges such as data quality and scalability persist, the ongoing refinement of efficient and generalizable fine-tuning methods offers considerable potential for broadening LLM applicability throughout various real-world tasks \citep{2310.12345}.

\input{timeline}
As shown in \textbf{Figure~\ref{fig:llm_timeline_vertical}}, the evolution timeline captures key milestones—from the initial surge of pre-training research in 2019, via the formal introduction of instruction tuning in 2021, to the release of foundational open source models (e.g., LLaMA) and the emergence of advanced tuning strategies such as LLaMA excitor and MDCure, thus contextualizing current challenges and directing future research trajectories in instruction-tuned LLMs.  

The development of scalable pipelines has emerged as a cornerstone in efforts to efficiently fine-tune large language models (LLMs). Recent studies have shown that instruction-based pipelines can dramatically enhance performance across benchmarks, with gains surpassing 75\% compared to pretrained baselines in multi-document tasks \citep{2410.23463}. These improvements originate from optimized data generation, efficient filtering approaches, and lightweight fine-tuning methods that preserve core capabilities while adapting models toward new objectives \citep{2404.00913}.

Accessibility and usability have become crucial priorities for developing training infrastructure. Frameworks have been introduced that enable fine-tuning with minimal configuration, facilitating wider adoption among practitioners lacking extensive technical expertise \citep{2308.10252}. Instruction-based approaches also improve usability by utilizing curated or synthetically produced prompts to guide fine-tuning in a task-aligned manner. Complementary methods, such as self-guided data selection and model compression via pruning, have been utilized to enhance both the training efficiency and generalization \citep{2308.12032, 2310.06694}.

Contrasting viewpoints have been offered concerning the relative advantages of supervised versus unsupervised fine-tuning methods. Certain studies highlight the benefits of small-scale, high-quality supervision \citep{2412.13337}, while others promote scalable pipelines utilizing synthetic or weakly labeled data \citep{2408.13296}. Likewise, parameter-efficient tuning techniques are often compared with resource-intensive methods that prioritize flexibility and general-purpose functionality \citep{2404.00913, 2410.23463}. These varied approaches reflect the increasing complexity of fine-tuning strategies and the necessity to align methodologies with deployment needs.

From a technical perspective, research continues to explore architectural and optimization innovations, such as gradient warm-up, learning rate scheduling, and mixed-precision fine-tuning. Benchmarking tools, including MMLU and MSCOCO, have been utilized to assess the transferability and generalization of fine-tuned models across tasks. Nevertheless, multiple challenges remain, such as limited scalability, resource constraints, data imbalances, and the requirement for robust optimization pipelines \citep{2412.13337, 2408.13296}. Emerging research directions indicate multi-modal adaptation, privacy-preserving training, and responsible model deployment.

Scalable LLM fine-tuning pipelines integrate data centric instruction methods with usability focused infrastructure and parameter efficient techniques. Ongoing progress will rely on balancing performance, accessibility, and alignment with particular application domains. As this field evolves, these pipelines are positioned to facilitate wider and more responsible LLM deployment in practical scenarios \citep{2410.23463}.

\section{Efficient Scaling, Model Selection, Compression, and Pruning of LLMs}

Efficient model selection and scaling approaches are crucial for deploying instruction-tuned large language models (LLMs) in resource-constrained environments. Recent research emphasizes the importance of comprehending model-scaling behaviors to inform system infrastructure design and resource allocation strategies \citep{2402.02314, 2406.03777, 2409.15156}. One proposed framework utilizes the concept of ``pre-learned data size'' for guiding model selection, achieving favorable trade-offs between performance and computational expense \citep{2402.02314}. This principle has demonstrated the ability to facilitate the identification of nearly optimal model configurations for fine-tuning, substantially decreasing training costs by multiple orders of magnitude in certain scenarios without significant performance loss.

\begin{table*}[htbp]
\centering
\caption{Scaling Laws and Optimization Strategies for Large Language Models}
\label{tab:scaling_laws}
\begin{tabular}{@{}lllll@{}}
\toprule
\textbf{Component} & \textbf{Scaling Law} & \textbf{Optimal Configuration} & \textbf{Complexity} & \textbf{Key Parameters} \\
\midrule
\multicolumn{5}{c}{\textit{Core Scaling Laws}} \\
\midrule
Model Size ($N$) & $L \propto N^{-0.076}$ & $N \propto C^{0.73}$ & $O(N^2)$ memory & $N \leq 10^{12}$ params \\
Data Scale ($D$) & $L \propto D^{-0.095}$ & $D \propto C^{0.27}$ & $O(D)$ processing & $D \leq 10^{13}$ tokens \\
Compute ($C$) & $L \propto C^{-0.050}$ & $C = 6ND$ (training) & $O(C)$ linear & FLOPs utilization \\
Joint Scaling & Chinchilla optimal & $N/D \approx 1/20$ & Pareto frontier & Budget allocation \\
\midrule
\multicolumn{5}{c}{\textit{Model Compression}} \\
\midrule
Structured Pruning & Sparsity-accuracy trade-off & $50\%-90\%$ sparsity & $O(N \log N)$ & Magnitude thresholding \\
Quantization & Precision reduction & INT8/INT4 & $O(N)$ & Calibration dataset \\
Knowledge Distillation & Teacher-student transfer & Student $10\%-30\%$ size & $O(N_T + N_S)$ & Temperature $\tau = 3-5$ \\
Low-Rank Adaptation & Parameter-efficient tuning & Rank $r = 8-256$ & $O(r \cdot d)$ & Task-specific adaptation \\
\midrule
\multicolumn{5}{c}{\textit{Architectural Innovations}} \\
\midrule
Sparse Attention & Attention pattern optimization & $10\%-30\%$ density & $O(N\sqrt{L})$ vs $O(NL^2)$ & Pattern design \\
Mixture of Experts & Conditional computation & Top-k routing, $k/E < 0.3$ & $O(kN)$, $k \ll E$ & Load balancing \\
Linear Attention & Linear complexity attention & RNN-like efficiency & $O(NL)$ & State representation \\
State Space Models & Efficient sequence modeling & Long-range dependencies & $O(NL)$ & Structured matrices \\
\midrule
\multicolumn{5}{c}{\textit{Data Efficiency}} \\
\midrule
Active Learning & Strategic data selection & $10\%-30\%$ selection ratio & $O(D \log D)$ & Uncertainty sampling \\
Curriculum Learning & Difficulty-ordered training & Progressive complexity & $O(D)$ & Pacing strategies \\
Data Augmentation & Synthetic data generation & $20\%-80\%$ augmentation & $O(G \cdot D)$ & Quality control \\
Few-Shot Learning & Sample-efficient adaptation & $k$-shot examples & $O(k)$ per task & Context design \\
\midrule
\multicolumn{5}{c}{\textit{System Optimization}} \\
\midrule
Data Parallelism & Gradient synchronization & $P = 8-1024$ GPUs & $O(P)$ communication & All-reduce bandwidth \\
Model Parallelism & Parameter sharding & Tensor/pipeline splits & $O(M/P)$ memory & Inter-device bandwidth \\
Pipeline Parallelism & Layer-wise distribution & $S = 2-16$ stages & $O(L/S)$ & Bubble minimization \\
Memory Optimization & Activation checkpointing & Time-memory trade-off & Recomputation overhead & Checkpoint granularity \\
\midrule
\multicolumn{5}{c}{\textit{Cost Analysis}} \\
\midrule
Training Cost & $C_{train} = 6ND$ & Linear in $N, D$ & GPU-hours scaling & Hardware utilization \\
Inference Cost & $C_{inf} = 2N \cdot \text{throughput}$ & Latency-throughput trade-off & Real-time constraints & Batch optimization \\
Storage Cost & Model size + checkpoints & Compression benefits & Linear in $N$ & Precision requirements \\
Total Ownership & Training + inference + storage & Lifetime optimization & Service duration & Amortization period \\
\midrule
\multicolumn{5}{l}{\footnotesize\textbf{Notation:} $L$ = loss function, $N$ = number of parameters, $D$ = dataset size (tokens),} \\
\multicolumn{5}{l}{\footnotesize$C$ = compute budget (FLOPs), $P$ = number of processors, $G$ = generation cost factor.} \\
\midrule
\multicolumn{5}{l}{\footnotesize\textbf{Key Relationships:}} \\
\multicolumn{5}{l}{\footnotesize$\bullet$ Optimal scaling: $N^{0.73} \cdot D^{0.27} \propto C$ (compute-optimal)} \\
\multicolumn{5}{l}{\footnotesize$\bullet$ Memory constraint: $N \times \text{precision} \leq \text{available memory}$} \\
\multicolumn{5}{l}{\footnotesize$\bullet$ Communication overhead: $\propto P \log P$ for all-reduce operations} \\
\multicolumn{5}{l}{\footnotesize$\bullet$ Quality threshold: Diminishing returns beyond high-quality data ceiling} \\
\bottomrule
\end{tabular}
\end{table*}

As shown in \textbf{Table~\ref{tab:scaling_laws}}, the scaling framework establishes fundamental power-law relationships where model performance scales as $N^{-0.076}$, $D^{-0.095}$, and $C^{-0.050}$ for parameters, data, and compute respectively, with the Chinchilla-optimal ratio $N/D \approx 1/20$ maximizing compute efficiency \citep{hoffmann2022training}. The framework encompasses model compression techniques achieving 50-90\% sparsity through structured pruning and INT8/INT4 quantization, architectural innovations including sparse attention reducing complexity from $\mathcal{O}(NL^2)$ to $\mathcal{O}(N\sqrt{L})$, and system optimizations addressing parallelization across $P = 8$-$1024$ processors \citep{fedus2022switch}. Cost analysis reveals training expenses scale as $C_{\text{train}} = 6ND$ while inference costs follow $C_{\text{inf}} = 2N \times \text{throughput}$, establishing quantitative guidelines for resource allocation in large-scale model development.

Empirical guidelines for deploying LLMs on edge devices have been introduced \citep{2406.03777}, highlighting the significance of optimal decisions between parameter learning and retrieval-augmented generation (RAG), along with the necessity for compression techniques. Moreover, research has challenged conventional principles in machine learning, promoting a transition toward scaling-centric methodologies \citep{2409.15156}. This paradigm shift is vital for understanding the limitations of traditional scaling approaches and creating new guiding principles for efficient model selection and scaling.

A recurring theme across these studies is the examination of scaling laws, which emphasizes their importance in strategic planning and system infrastructure development \citep{2208.08489, 2412.06540, 2412.01505}. The significance of resource limitations is another frequent topic, with numerous studies addressing the challenges of deploying LLMs within resource-limited environments, such as edge devices \citep{2405.03146, 2410.16208, 2407.17467}. In these resource-constrained situations, prompt engineering methods, as shown in Table~\ref{tab:prompt_engineering},  provide parameter-efficient alternatives to complete model fine-tuning, allowing enhanced performance without altering model weights. Model selection and comparison are equally important, with multiple studies concentrating on selecting the most appropriate model considering resource constraints and comparing models at scale \citep{2312.03732, 2501.06802, 2412.04315}.

The trade-offs among model size, precision, and performance have been thoroughly investigated, providing insights into the effects of scaling strategies \citep{2405.15319, 2409.03444, 2501.02068}. While larger models typically outperform their smaller counterparts and demonstrate greater resilience to precision reduction \citep{2405.03146}, parameter scaling faces inherent limitations, making data scaling a more promising direction for continued performance improvement \citep{2208.08489}. Consequently, compression techniques and quantization methods have been extensively evaluated as complementary approaches for reducing memory requirements while maintaining the effectiveness of the model \citep{2305.09246, 2306.09479}.

Research has formalized these relationships through empirical scaling laws that characterize the power-law dependence of the model performance on the key factors as follows:
\begin{equation}
L(N, D) = \frac{A}{N^{\alpha}} + \frac{B}{D^{\beta}} + E,
\label{eq:scaling_law}
\end{equation}
where $L$ represents the validation loss, $N$ denotes the number of model parameters, $D$ is the dataset size (typically measured in tokens), and $A$, $B$, $\alpha$, $\beta$, $E$ are empirically determined constants. The exponents $\alpha$ and $\beta$ typically range from 0.05 to 0.1, indicating diminishing returns with an increased scale. This formulation enables practitioners to predict the model performance given the computational budgets and guides optimal resource allocation between the model size and training data volume.

The scaling law reveals that both the model parameters and data contribute independently to performance improvement, but with different efficiency profiles. When computational resources are constrained, the law suggests prioritizing data scaling over model scaling for better performance gains per unit cost, particularly in the over-parameterized regime, where models have sufficient capacity but lack adequate training data.

Technical advances in scaling laws and experimental evaluations provide a deeper understanding of the relationships among model size, data volume, and performance \citep{2402.02314, 2406.03777}. However, significant limitations and challenges remain, including resource constraints, scalability bottlenecks, and an incomplete understanding of scaling principles \citep{2409.15156, 2410.16208, 2501.06802}. Addressing these challenges is crucial for developing efficient model selection and scaling techniques that enable the effective deployment of instruction-tuned LLMs in resource-constrained environments.

Recent research has achieved substantial progress in understanding the importance of scaling laws and efficient model selection for deploying instruction-tuned LLMs in practical applications. The development of scaling laws and empirical guidelines for edge devices has provided valuable insights into the challenges and constraints of traditional scaling techniques. As this field continues to evolve, it is essential to address outstanding challenges and limitations, investigate alternative strategies such as data scaling, and thoroughly evaluate techniques, including compression and quantization. Through these efforts, researchers can develop more efficient system infrastructure and strategic frameworks, ultimately facilitating the effective deployment of instruction-tuned LLMs in resource-constrained settings \citep{2402.02314, 2406.03777, 2409.15156}.

Model compression and pruning are crucial strategies for addressing the challenges related to the expanding scale of large language models (LLMs), particularly in computationally constrained environments. Recent studies have presented pruning frameworks that reduce parameter counts while maintaining performance, enabling the feasible deployment of LLMs on resource-limited hardware \citep{2402.05406, 2310.05015}.

These methods typically optimize an objective function that balances task performance and model sparsity:
\begin{equation}
\min_{\mathbf{m}} \mathcal{L}(\mathbf{\theta} \odot \mathbf{m}) + \lambda \|\mathbf{1} - \mathbf{m}\|_0,
\label{eq:pruning_obj}
\end{equation}
where $\mathbf{\theta}$ represents the original model parameters, $\mathbf{m} \in \{0,1\}^{|\theta|}$ is a binary mask (0 for pruned parameters, 1 for retained), $\odot$ denotes element-wise multiplication, $\mathcal{L}$ is the task-specific loss function, $\|\mathbf{1} - \mathbf{m}\|_0$ counts the number of pruned parameters, and $\lambda > 0$ controls the sparsity-performance trade-off.

Structured pruning methods are particularly effective for this purpose. By identifying and eliminating entire parameter groups, such as neurons, attention heads, or layers, structured approaches enable more hardware-compatible implementations while preserving the essential model functionality \citep{2310.05015, 2406.15734}. This type of pruning results in significant reductions in memory consumption and inference latency, making it more suitable for edge and real-time applications than unstructured pruning techniques.

A common approach to identify important attention heads for retention uses gradient-based importance scoring:
\begin{align}
\text{importance}(h_{i,j}) &= \mathbb{E}_{x \sim \mathcal{D}} \left[ \left| \frac{\partial \mathcal{L}(x)}{\partial h_{i,j}} \right| \cdot |h_{i,j}| \right], \label{eq:head_importance_a}\\
\text{rank} &= \text{argsort}(\{\text{importance}(h_{i,j})\}),
\label{eq:head_importance_b}
\end{align}
where $h_{i,j}$ represents the output of the $j$-th attention head in the $i$-th layer, and the importance score combines gradient magnitude with activation magnitude, averaged over dataset $\mathcal{D}$ \citep{michel-etal-2019-sixteen}. Heads with lower importance scores are candidates for removal.

Beyond general-purpose methods, domain-adaptive pruning strategies have been explored. These approaches utilize task-specific information to remove parameters that are less relevant to particular domains while maintaining general capabilities \citep{2405.06275}. These techniques provide promising directions for creating compact and flexible LLM variants tailored for specialized applications, achieving domain-specific efficiency without sacrificing cross-domain transferability.

An analysis of recent research on model compression and pruning reveals multiple recurring themes, including the vital importance of efficient parameter reduction, the effectiveness of pruning in minimizing model dimensions, and the increasing adoption of structured pruning approaches. Studies have highlighted the benefits of domain-specific pruning methods, which enable task-relevant parameter selection and enhanced deployment efficiency within specialized contexts. For example, gradient-free and gradient-based techniques have both shown strong performance in reducing model size while maintaining accuracy \citep{2402.05406, 2310.05015}, and adaptive pruning frameworks have been created for domain-specific situations \citep{2405.06275}.

The methodologies utilized in these studies are remarkably diverse, encompassing algorithmic innovations and evaluation protocols customized for various application constraints. Certain approaches present multi-stage pruning schemes that dynamically eliminate redundant parameters without backpropagation \citep{2405.06275}, while others integrate cost models to assess compression trade-offs in offline and online learning contexts \citep{2406.15734}. The evaluation metrics ranged from classification accuracy and F1-score to perplexity, demonstrating the heterogeneity of the use cases. Additionally, survey efforts have synthesized findings regarding the current limitations and emerging trends in model compression research \citep{2308.07633, 2402.01799}.

Despite substantial progress, model pruning is associated with multiple challenges. These challenges include the high computational overhead associated with pruning large-scale models, potential performance degradation from aggressive parameter removal, and the complexity of developing domain-adaptive pruning pipelines. To address these issues, recent research has investigated more efficient optimization procedures and introduced novel evaluation strategies that better capture performance trade-offs \citep{2410.17711, 2312.10793}. Nevertheless, balancing efficiency and model fidelity remains a primary concern.

The implications of advances in compression and pruning extend beyond LLMs and are applicable to domains such as natural language processing, computer vision, and speech recognition. As models continue to grow in scale, developing lightweight and effective deployment approaches has become increasingly critical. Building upon insights from recent research, future investigations might combine model compression with complementary methods, such as knowledge distillation \citep{2501.00123} and transfer learning \citep{2502.00234}, promoting cross-domain collaboration and innovation.

\section{Advanced Techniques and Technical Challenges in LLM Development}
Advanced methods for enhancing large language model (LLM) performance have been thoroughly investigated in recent studies, focusing on improving instruction-following capabilities through three main approaches: architectural enhancements, data-centric augmentation, and neuro-symbolic reasoning integration. 

\noindent \textbf{Architectural Optimization.} The LLaMA-Excitor \citep{2404.00913} demonstrates architectural optimization by improving attention allocation within models through indirect feature interaction, achieving a +6\% improvement on the MMLU benchmark by enhancing the attention mechanism:
\begin{equation}
\mathbf{A}^l = \text{softmax}\left(\frac{\mathbf{Q}^l(\mathbf{K}^l)^T}{\sqrt{d}} + \mathbf{E}^l\right)\mathbf{V}^l,
\label{eq:excitor_attention}
\end{equation}
where $\mathbf{E}^l \in \mathbb{R}^{n \times n}$ represents the learnable excitation matrix that enhances instruction tracking for the $l$-th layer, and $n$ denotes the sequence length.

\noindent \textbf{Neuro-symbolic Integration.} Complementing architectural approaches, neuro-symbolic methods \citep{2407.18722} integrate structured symbolic reasoning with neural components through a comprehensive pipeline involving: (1) symbolic task planners for instruction decomposition, (2) neural semantic parsers for grounding, and (3) neuro-symbolic executors for action specification. These methods demonstrate improved performance on compositional and structured reasoning tasks while maintaining the flexibility of the neural architecture.

\noindent \textbf{Data Augmentation Strategies.} Data augmentation is equally important for instruction tuning effectiveness. Novel approaches, such as Mosaic-IT \citep{2405.13326}, present compositional frameworks that synthesize new instruction-response pairs entirely from existing datasets without human annotations or external teacher models. 

Given a base dataset $\mathcal{D}_{\text{base}} = \{(I_i, R_i)\}_{i=1}^N$, this technique generates new samples through a controlled mixing
\begin{align}
\mathcal{D}_{\text{mix}} &= \{(I_{\text{mix}}, R_{\text{mix}})\}, \label{eq:mosaic_a}\\
\text{where } I_{\text{mix}} &= f_{\text{comp}}(I_i, I_j), \label{eq:mosaic_b}\\
R_{\text{mix}} &= g_{\text{comp}}(R_i, R_j), \label{eq:mosaic_c}
\end{align}
and $f_{\text{comp}}, g_{\text{comp}}$ denote composition functions such as concatenation, interleaving, or instruction rewriting, achieving consistent improvements across benchmarks with up to 80\% reduction in training cost.

In contrast, IDEA-MCTS \citep{2410.10392} formulates instruction synthesis as a tree-based search problem, leveraging Monte Carlo Tree Search (MCTS) to explore instruction candidate spaces guided by reward functions that evaluate informativeness, diversity, and executability. This approach enables scalable instruction generation, particularly for multi-step and complex reasoning tasks that require systematic exploration of solution spaces.

\noindent \textbf{Multi-step Instruction Processing.} Multi-step instruction following presents significant challenges, necessitating reasoning across multiple information sources and producing coherent responses throughout documents or tasks. It requires hierarchical processing that integrates semantic enrichment (e.g., named entity recognition and dependency parsing) with multi-document reasoning methods, such as knowledge graph traversal and logical inference. These systems typically comprise three core components: (1) preprocessing documents into structured representations, (2) conducting intermediate cross-document reasoning, and (3) generating responses using instruction-tuned decoders. This hierarchical architecture enhances instruction alignment and improves model performance on compositional, domain-specific, and instruction-intensive tasks that require a comprehensive understanding of inter-document relationships and multi-step reasoning capabilities.

\begin{figure*}[t]
\centering
\includegraphics[width=\textwidth]{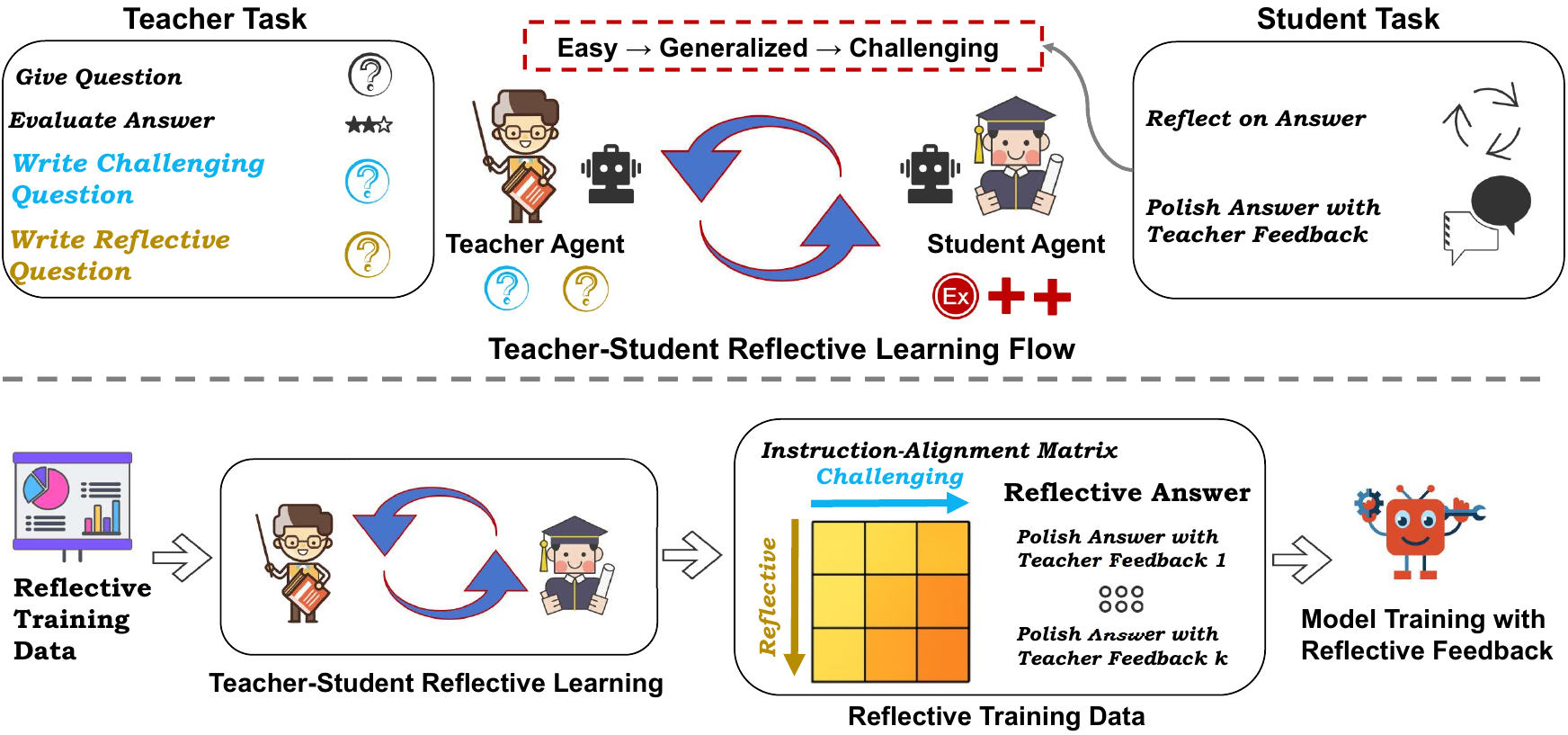}
\caption{Teacher-Student Reflective Learning Flow and Instruction-Alignment Matrix. The diagram illustrate the iterative learning process among teacher and student agents, with feedback cycles and instruction alignment for reflective learning.}
\label{fig:reflective_learning}
\end{figure*}

\begin{algorithm}[t]
\caption{Multi-Step Instruction Following with Reflective Learning}
\label{alg:multi_step_reflective}
\small
\begin{algorithmic}[1]
\REQUIRE Document set $D = \{d_1, d_2, \ldots, d_n\}$, Instruction $I$, Difficulty levels $\mathcal{L} = \{l_1, l_2, \ldots, l_k\}$, Max iterations $T$
\ENSURE Final response $R_{\text{final}}$, Alignment matrix $\mathbf{A} \in \mathbb{R}^{k \times T}$
\STATE \textbf{Initialize:} $t \leftarrow 0$, $\mathbf{A} \leftarrow \mathbf{0}_{k \times T}$, $\text{converged} \leftarrow \text{False}$
\STATE \textbf{Document Processing:}
\FOR{each $d_i \in D$}
\STATE $d_i^{\text{proc}} \leftarrow \textsc{Preprocess}(d_i)$
\STATE $S_i \leftarrow \textsc{Retrieve}(d_i^{\text{proc}}, I)$
\ENDFOR
\STATE \textbf{Initial Response Generation:}
\STATE $C_0 \leftarrow \textsc{Aggregate}(\{S_i\}_{i=1}^n)$
\STATE $R_0 \leftarrow \textsc{Generate}(C_0, I)$
\STATE \textbf{Reflective Learning Loop:}
\WHILE{$\neg\text{converged}$ \AND $t < T$}
\FOR{each difficulty level $l_j \in \mathcal{L}$}
\STATE $Q_{j,t} \leftarrow \textsc{TeacherAgent.GenerateQuestion}(I, l_j, R_t)$
\STATE $R_{j,t} \leftarrow \textsc{StudentAgent.Answer}(Q_{j,t}, C_t)$
\STATE $F_{j,t} \leftarrow \textsc{TeacherAgent.Evaluate}(R_{j,t}, Q_{j,t})$
\STATE $\mathbf{A}[j,t] \leftarrow \textsc{ComputeAlignment}(R_{j,t}, I, l_j)$
\ENDFOR
\STATE $R_{t+1} \leftarrow \textsc{StudentAgent.Reflect}(R_t, \{F_{j,t}\}_{j=1}^k)$
\STATE $\text{converged} \leftarrow \textsc{CheckConvergence}(\mathbf{A}, t, \epsilon)$
\STATE $t \leftarrow t + 1$
\ENDWHILE
\RETURN $R_{\text{final}} = R_t$, $\mathbf{A}$
\end{algorithmic}
\end{algorithm}

Teacher-student reflective learning paradigms have transformed instruction tuning by introducing internal feedback mechanisms that enable models to continually refine their outputs via iterative self-improvement cycles, as shown in \textbf{Figure~\ref{fig:reflective_learning}}. This approach fundamentally differs from traditional data augmentation techniques by emphasizing internal reflection instead of external synthetic data creation, utilizing a systematic framework where teacher agents present increasingly difficult questions to student agents, with students reflecting upon their responses and improving them through structured feedback loops (see \textbf{Algorithm~\ref{alg:multi_step_reflective}}). 

The instruction-alignment matrix $\mathbf{A} \in \mathbb{R}^{k \times T}$ captures this evolutionary process, where each element represents the alignment score for difficulty level $l_j$ at iteration $t$:
\begin{equation}
\label{eq:alignment_matrix}
\mathbf{A}[j,t] = \text{sim}(\text{embed}(R_{j,t}), \text{embed}(I)) \cdot \text{quality}(R_{j,t}, l_j),
\end{equation}
\begin{equation}
\label{eq:convergence}
\text{convergence} = \|\mathbf{A}[:, t] - \mathbf{A}[:, t-1]\|_2 < \epsilon,
\end{equation}
where $\epsilon > 0$ is the predefined convergence threshold.

This formulation combines the semantic similarity between the response and instruction embeddings with response appropriateness metrics. It enables systematic tracking of learning progress across varying difficulty levels, providing quantitative measures of instruction-following improvement. To determine when the student model sufficiently refines its responses, a convergence criterion is applied based on the change in alignment scores across successive iterations, as shown in \textbf{Equation~\ref{eq:convergence}}. This condition ensures that the iterative refinement process halts when the improvement in alignment becomes marginal, with more challenging tasks typically requiring additional iterations to meet the convergence threshold and achieve optimal instruction adherence.

\begin{figure}[t]
\centering
\begin{tikzpicture}[scale=0.8]
   
    \foreach \i in {0,1,2,3} {
        \foreach \j in {0,1,2,3,4} {
            \pgfmathsetmacro{\intensity}{15 + \i*10 + \j*8}
            \fill[orange!\intensity] (\j*1.4,3.5-\i*1.0) rectangle ++(1.2,0.9);
            \draw[thick] (\j*1.4,3.5-\i*1.0) rectangle ++(1.2,0.9);
        }
    }

    \node[left, font=\small] at (-0.8,4.05) {Easy};
    \node[left, font=\small] at (-0.6,3.05) {Medium};
    \node[left, font=\small] at (-0.8,2.05) {Hard};
    \node[left, font=\small] at (-0.8,1.05) {Expert};
    
  \node[below, font=\small] at (0.7,0.1) {$t_1$};
    \node[below, font=\small] at (2.1,0.1) {$t_2$};
    \node[below, font=\small] at (3.5,0.1) {$t_3$};
    \node[below, font=\small] at (4.9,0.1) {$t_4$};
    \node[below, font=\small] at (6.3,0.1) {$t_5$};

  \draw[->, thick] (-0.7,2.5) -- (-0.7,0.3);
\node[rotate=90,font=\small] at (-0.35,1.4) {Difficulty};
    \draw[->, thick] (3.0,-0.6) -- (5.5,-0.6) node[midway,below,font=\small] {Iterations};

    \begin{scope}[shift={(7.5,1.5)}]
        \fill[orange!15] (0,0) rectangle (0.4,0.3);
        \draw (0,0) rectangle (0.4,0.3);
        \node[right, font=\tiny] at (0.5,0.15) {Low};
        
        \fill[orange!35] (0,0.4) rectangle (0.4,0.7);
        \draw (0,0.4) rectangle (0.4,0.7);
        
        \fill[orange!55] (0,0.8) rectangle (0.4,1.1);
        \draw (0,0.8) rectangle (0.4,1.1);
        \node[right, font=\tiny] at (0.5,0.95) {High};
        
        \node[above, font=\scriptsize] at (0.2,1.3) {Alignment};
    \end{scope}

    \node[below, font=\footnotesize] at (3.5,-1.2) {$\mathbf{A} \in \mathbb{R}^{4 \times 5}$};
    
\end{tikzpicture}
\caption{Instruction-Alignment Matrix Evolution. The matrix shows progressive improvement in alignment scores throughout the difficulty levels (rows) and training iterations (columns). Darker orange shading indicate higher alignment quality, with expert-level tasks needing additional iterations for achieving optimal instruction adherence.}
\label{fig:alignment_matrix}
\end{figure}

The integration of these advanced methodologies, including attention modulation, structured planning, tree search-based optimization, and reflective learning frameworks, has shown substantial improvements in instruction-following performance and exposed important directions for future progress. Recent scalable fine-tuning pipelines demonstrate particular promise for enhancing instruction-following abilities without depending on human-annotated data \citep{2410.23463}. These pipelines focus on addressing content fragmentation throughout documents and enhancing model performance on standardized benchmarks via sophisticated multi-document modeling and symbolic reasoning integration. The matrix visualization in \textbf{Figure~\ref{fig:alignment_matrix}} illustrates the practical effectiveness of these methods, demonstrating how alignment scores gradually improve across iterations. More difficult complexity levels require extra refinement cycles to achieve optimal instruction compliance, thus offering a quantitative framework for assessing and optimizing instruction-following systems throughout varied task complexities and application areas.

\begin{figure*}[t]
    \centering
    \includegraphics[width=\linewidth]{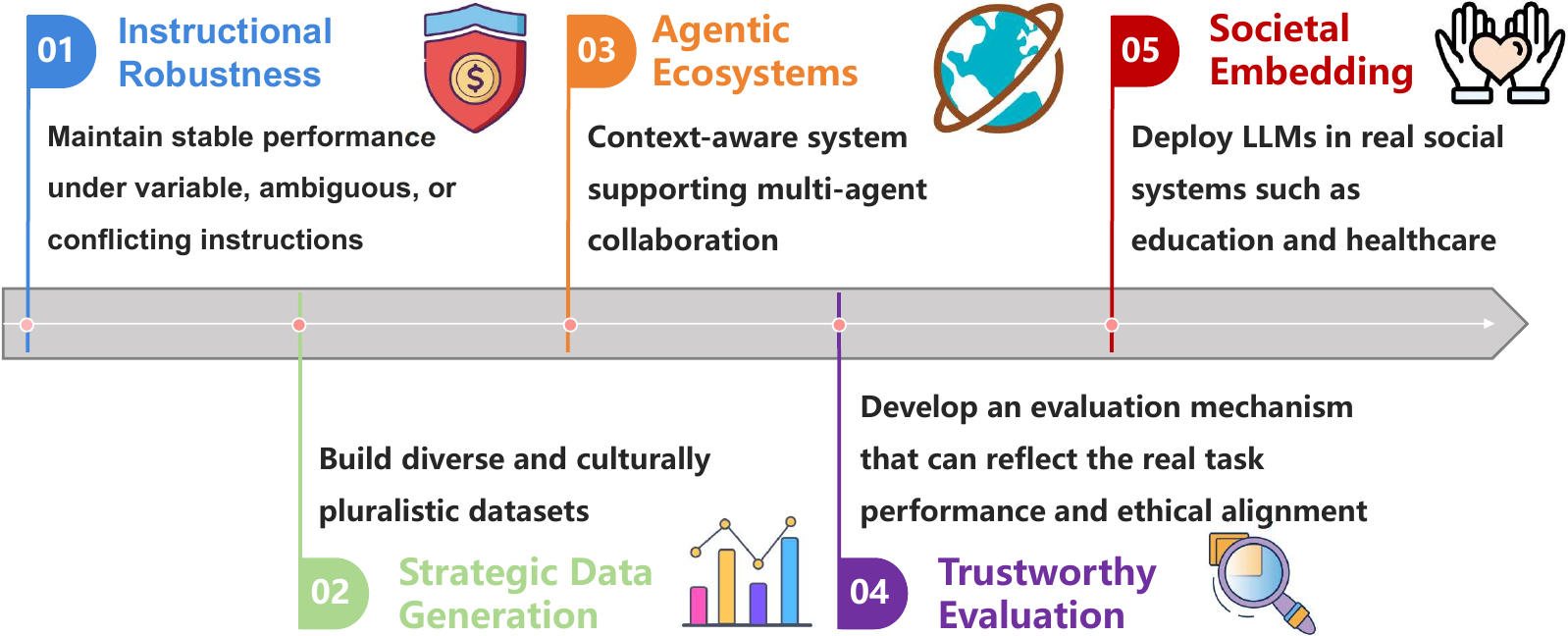}
    \caption{Future research roadmap for instruction-tuned LLMs throughout five pillars: Instructional Robustness, Strategic Data Generation, Agentic Ecosystems, Trustworthy Evaluation, and Societal Embedding.}
    \label{fig:future_outlook}
\end{figure*}

\section{Evaluating LLM Performance: Benchmarks, Metrics, and Future Directions}

The evaluation of instruction-tuned large language models (LLMs) has become increasingly complex and crucial as a research focus. Recent research has presented multilingual and instruction-specific evaluation frameworks designed to capture performance across languages, tasks, and cultural contexts \citep{2312.07398, 2502.04688, 2410.12972}. Together, these efforts highlight the necessity of evaluation approaches that align with practical deployment situations and varied user needs.

Instruction fine-tuning has been demonstrated to enhance zero-shot generalization; however, it introduces challenges in evaluation, especially regarding how models follow instructions in zero-shot contexts \citep{2310.14103, 2410.12972}. These challenges have resulted in the adoption of more detailed, context-aware metrics, such as LLM-generated evaluation scores, which better capture alignment and response quality \citep{2310.14103}. Furthermore, multilingual benchmarks now include languages such as French, Japanese, and Spanish, revealing performance disparities and the requirement for culturally aware evaluation \citep{2502.04688}. While certain studies concentrate on developing new datasets for instruction-following and multilingual assessment \citep{2312.07398, 2410.12972}, others reinterpret evaluation results utilizing advanced statistical modeling \citep{2403.15250}, demonstrating the continuing debate among general-purpose model quality and task-specific capabilities. The interaction between innovative metrics and dataset design continues to shape the changing landscape of LLM evaluation.

Despite these advances, challenges persist, including assessing reasoning within low-resource contexts, standardizing multilingual protocols, and aligning benchmark metrics with practical user expectations. Current evaluation approaches frequently overlook the balance among performance, robustness, and cultural sensitivity concerns. Addressing these problems, future research needs to focus on creating transparent, adaptable, and context-sensitive evaluation frameworks to ensure that instruction-tuned LLMs are accurate, dependable, and properly reflect the varied global environments in which they operate.

\section{The Evolving Landscape of Large Language Models and Their Applications}

The development trajectory of large language models (LLMs) has shifted from a scale-centric paradigm to one emphasizing controllability, alignment, and computational efficiency. A survey of more than 5,000 publications from 2017 to 2023 demonstrates the scope of research topics covering foundational architectures, instruction tuning, and application-specific adaptations \citep{2304.02020}. This transition reflects the increasing acknowledgment of the versatility and fragility of current LLM systems.

Instruction tuning has emerged as a central approach for aligning LLM outputs with human intentions across various domains, such as education \citep{2403.18105}, healthcare, engineering, and humanities \citep{2502.05467}. While instruction tuning has shown measurable advantages, research has expressed concerns about knowledge erosion \citep{2402.05119} and instruction ambiguity, especially in zero-shot and multilingual situations \citep{2410.12972}. The dynamic relationships between in-context learning and instruction tuning remain an active research area, with a growing focus on efficient fine-tuning, multi-agent coordination, and neuro-symbolic integration \citep{2404.00913, 2407.18722}.

Ongoing challenges continue to shape the landscape of LLM research. Problems such as interpretability, resource demands, and cultural bias within moral reasoning \citep{2412.19926} emphasize the necessity for increased transparent model behavior and culturally inclusive evaluation standards. In response, researchers have promoted value-sensitive design principles and training protocols that explicitly integrate ethical alignment goals \citep{2406.05392, 2410.19812}.

As this field moves beyond raw scale as the main driver of performance, future advances are anticipated in architectural innovations, data-centric tuning approaches, cross-modal reasoning, and adaptive instruction processing. Maintaining this progress will require balancing technical developments with thoughtful consideration of societal effects, ensuring that LLMs evolve not only in terms of capability but also in terms of safety, accountability, and widespread usefulness.

\section{Open Challenges and Future Research Directions in LLMs: Ethics, Explainability, and Societal Impact}

Instruction-tuned LLMs present major challenges at the intersection of ethics, explainability, and social responsibility. As their deployment expands throughout sensitive and high-impact areas, researchers have emphasized the significance of aligning model behavior with ethical principles and social norms \citep{2410.19812}. Primary concerns include ensuring factual accuracy, reducing learned biases, preventing copyright violations, and maintaining cultural sensitivity within both training data and outputs \citep{2412.19926, 2406.05392}. Instead of treating ethics as a secondary concern, these studies promote its integration across the model development and evaluation lifecycle. Considering the expanding role of LLMs in influencing information access, assisting decisions, and mediating human-computer interactions, addressing their ethical and societal implications is essential for ensuring that such systems are transparent, trustworthy, and aligned with wider human values.

As illustrated in \textbf{Figure~\ref{fig:future_outlook}}, the next phase of instruction-tuning research will revolve around five interlinked pillars. These include maintaining robustness under ambiguous or conflicting instructions, building culturally pluralistic datasets, enabling context-aware multi-agent collaboration, establishing evaluation frameworks that reflect both performance and ethical alignment, and embedding LLMs into real-world domains, such as education and healthcare.

A central theme emerging from recent research is the necessity of value alignment. Certain studies support explicit moral alignment objectives \citep{2310.07251}, while others stress the development of generalizable ethical reasoning frameworks addressing value pluralism within global contexts. Nevertheless, empirical findings show that larger models frequently display preferences toward specific moral paradigms, such as deontological reasoning, thus demonstrating the complexity of handling ethical dilemmas across varied social environments \citep{2412.19926}.

The societal implications of LLM deployment are expanding, especially in domains such as education and scientific communication \citep{2406.14871, 2306.09928}. Recent critiques have highlighted insufficient methodological rigor and transparency within current development pipelines \citep{2408.15409}. Trends, including inconsistent inclusion of ethical disclaimers and decreasing attention to emergent behaviors, indicate a requirement for more reproducible, ethically based research practices.

Technical factors, such as dataset construction and model architecture, play important roles in addressing these challenges. Efforts to diversify training corpora aim to reduce overfitting toward dominant Western viewpoints \citep{2406.05392, 2412.19926}, while architectural innovations investigate better handling of ethically sensitive tasks \citep{2404.06404}. However, ongoing limitations, including cultural bias and inadequate standardization, remain substantial barriers to achieving robust and fair model behavior.

Future research should prioritize the explainability and transparency of LLMs, involving a deeper examination of internal decision-making processes \citep{2401.12874} and the creation of robust evaluation protocols for ethical reasoning \citep{2406.00936}. Additionally, incorporating varied datasets that reflect global moral perspectives is crucial. Human-in-the-loop evaluation frameworks and feedback-based refinement are equally significant for enhancing ethical alignment within AI models.

Addressing these ethical, explainability, and societal challenges requires interdisciplinary collaboration among computer scientists, ethicists, philosophers, and sociologists. By integrating these priorities within the instruction-tuning lifecycle, we can create LLMs that are not only technically capable but also transparent, equitable, and aligned with the varied requirements of global society \citep{2408.15409}.

\section{Conclusion}

This survey examines instruction tuning as a critical paradigm for aligning large language models with human intentions and domain-specific requirements. Through systematic analysis of the complete pipeline—spanning manual annotation, distillation-based generation, and self-improvement mechanisms for dataset construction, alongside full-parameter and parameter-efficient fine-tuning strategies—we reveal significant trade-offs between quality, scalability, and computational efficiency. While instruction tuning demonstrates substantial effectiveness in enhancing model alignment, particularly in high-stakes applications, our evaluation exposes notable gaps in current benchmarks regarding faithfulness, fairness, and robustness assessment. These limitations underscore the urgent need for more comprehensive evaluation frameworks that capture the complexity of real-world deployment scenarios. Moving forward, the field must transcend the traditional focus on scale and prioritize the development of reliable, interpretable, and ethically aligned systems. Success in this endeavor requires integrating rigorous data curation with human-centered evaluation protocols, ensuring that future instruction-tuned models not only excel in capability but also embody the trustworthiness and value alignment essential for responsible AI deployment across diverse global contexts.

\bibliographystyle{apalike}
\bibliography{cite}

\clearpage
\appendices
\twocolumn[{%
  \centering
  {\LARGE\bfseries Appendix\par}
  \vspace{5mm}
  \begin{minipage}{0.92\textwidth}
    \small
    This appendix provides supplementary materials to complement the main text. 
    It includes a taxonomy of instruction-tuned LLMs (Figure~\ref{fig:instruction_tuned_llm_taxonomy}), 
    two comprehensive comparison tables covering commercial and open source models, efficient fine-tuning methods, 
    multi-document approaches, and domain-specific models, 
    as well as a summary table of prompt-engineering templates. 
    Together, these resources serve as detailed references for researchers and practitioners.
  \end{minipage}
}]
\vspace*{-0.25\baselineskip}

\input{forest}
\input{comp_table1}
\input{comp_table2}
\input{prompt_engineering_table}

\end{document}

%% file: timeline.tex
\begin{figure}[!t]
\centering
\begin{tikzpicture}[
  timeline/.style={draw=gray!60, very thick, -Stealth},
  event/.style={circle, draw=black, fill=blue!20, minimum size=6mm, inner sep=1pt},
  major_event/.style={circle, draw=black, fill=red!30, minimum size=7mm, inner sep=1pt},
  label/.style={font=\footnotesize, align=left, text width=3.5cm},
  description/.style={font=\scriptsize, text width=3.5cm, align=left, text=gray!80},
  year/.style={font=\footnotesize, align=right, text width=1cm}
]
\def\yearSpace{3.0cm}
\def\minEventSpace{1.5cm}

\def\posFirst{0}
\def\posSecond{-3.0cm}
\def\posThird{-6.0cm}
\def\posInstructGPT{-7.5cm}
\def\posFourth{-9.0cm}
\def\posSheared{-10.5cm}
\def\posFifth{-12.0cm}
\def\posExcitor{-13.5cm}
\def\posSixth{-15.0cm}

\draw[timeline] (0,\posFirst) -- (0,\posSixth-2cm);

\foreach \y/\year in {
  \posFirst/2019,
  \posSecond/2020,
  \posThird/2021,
  \posFourth/2022,
  \posFifth/2023,
  \posSixth/2024
} {
  \draw[gray!40] (-0.2,\y) -- (0.2,\y);
  \node[year, left=1.5cm] at (0,\y) {\year};
}

\draw[line width=5pt, blue!20] (-0.8,\posFirst) -- (-0.8,\posSecond);
\draw[line width=5pt, green!30] (-0.8,\posSecond) -- (-0.8,\posThird);
\draw[line width=5pt, orange!30] (-0.8,\posThird) -- (-0.8,\posFifth);
\draw[line width=5pt, red!20] (-0.8,\posFifth) -- (-0.8,\posSixth-2cm);

\foreach \y in {\posFirst,\posSecond,\posThird,\posInstructGPT,\posFourth,\posSheared,\posFifth,\posExcitor,\posSixth} {
  \draw[gray!40, dashed, thin] (-0.8,\y) -- (0,\y);
}

\node[major_event] (e1) at (0,\posFirst) {};
\node[label, right=0.3cm] at (0,\posFirst) {\textbf{Early LLMs}};
\node[description, right=0.3cm] at (0,\posFirst-0.7cm) {Pretraining language models gain attention};

\node[event] (e2) at (0,\posSecond) {};
\node[label, right=0.3cm] at (0,\posSecond) {\textbf{Pre-instruction LLMs}};
\node[description, right=0.3cm] at (0,\posSecond-0.7cm) {GPT-3 demonstrates powerful language abilities};

\node[major_event] (e3) at (0,\posThird) {};
\node[label, right=0.3cm] at (0,\posThird) {\textbf{Instruction Tuning Emerges}};
\node[description, right=0.3cm] at (0,\posThird-0.7cm) {Introduction of instruction tuning concept};

\node[event] (e4) at (0,\posInstructGPT) {};
\node[label, right=0.3cm] at (0,\posInstructGPT) {\textbf{InstructGPT}};
\node[description, right=0.3cm] at (0,\posInstructGPT-0.7cm) {OpenAI's first systematic instruction-tuned LLM};

\node[major_event] (e5) at (0,\posFourth) {};
\node[label, right=0.3cm] at (0,\posFourth) {\textbf{LLaMA}};
\node[description, right=0.3cm] at (0,\posFourth-0.7cm) {Meta releases open-source large language model};

\node[event] (e6) at (0,\posSheared) {};
\node[label, right=0.3cm] at (0,\posSheared) {\textbf{Sheared LLaMA}};
\node[description, right=0.3cm] at (0,\posSheared-0.7cm) {Structured pruning for instruction-tuned models};

\node[event] (e7) at (0,\posFifth) {};
\node[label, right=0.3cm] at (0,\posFifth) {\textbf{DoG-Instruct}};
\node[description, right=0.3cm] at (0,\posFifth-0.7cm) {Text-grounded instruction wrapping approach};

\node[major_event] (e8) at (0,\posExcitor) {};
\node[label, right=0.3cm] at (0,\posExcitor) {\textbf{LLaMA-Excitor}};
\node[description, right=0.3cm] at (0,\posExcitor-0.7cm) {Indirect feature interaction for instruction tuning};

\node[event] (e9) at (0,\posSixth) {};
\node[label, right=0.3cm] at (0,\posSixth) {\textbf{MDCure}};
\node[description, right=0.3cm] at (0,\posSixth-0.7cm) {Scalable multi-document instruction pipeline};

\node[font=\small\bfseries, align=center] at (2.5,\posFirst+1cm) {Evolution of Instruction-Tuned LLMs};

\node[text width=5cm, align=center, font=\scriptsize] at (3.5,\posSixth-2cm) 
{Timeline of Instruction-Tuned Large Language Models: Key milestones from early models to advanced instruction tuning techniques like LLaMA-Excitor and MDCure.};

\node[left, rotate=90] at (-0.8,-0.5cm) {\small Early Models};
\node[left, rotate=90] at (-0.8,-3.2cm) {\small Pre-Instruction Era};
\node[left, rotate=90] at (-0.8,-6.8cm) {\small Instruction Tuning Development};
\node[left, rotate=90] at (-0.8,-12.5cm) {\small Advanced Instruction Tuning};

\end{tikzpicture}

\caption{Evolution of Instruction-Tuned Large Language Models from 2019 to 2024, showing major developments and technological advancements in the field. Red nodes indicate transformative milestones (major events) that fundamentally changed the LLM landscape, such as the emergence of instruction tuning and release of foundational open-source models. Blue nodes represent important but incremental advancements in model development or techniques. The timeline illustrates progression through four distinct phases: Early Models (blue band), Pre-Instruction Era (green band), Instruction Tuning Development (orange band), and Advanced Instruction Tuning (red band).}

\label{fig:llm_timeline_vertical}
\end{figure}

%% file: forest.tex
\definecolor{oceanblue}{RGB}{230, 245, 255}      
\definecolor{forestgreen}{RGB}{240, 255, 245}   
\definecolor{sunsetorange}{RGB}{255, 245, 235}  
\definecolor{rosepink}{RGB}{255, 240, 250}      
\definecolor{goldenyellow}{RGB}{255, 250, 230}  
\definecolor{skyblue}{RGB}{235, 250, 255}       
\definecolor{lavenderblue}{RGB}{245, 240, 255}  

\begin{figure*}[htbp]
    \vspace{-1mm}
    \centering
    \resizebox{0.95\textwidth}{!}{
        \begin{forest}
            for tree={
                child anchor=west,
                parent anchor=east,
                grow'=east,
                anchor=west,
                base=left,
                font=\normalsize,
                rectangle,
                draw=black,
                rounded corners,
                align=left,
                minimum width=5em,
                edge+={darkgray, line width=1.2pt},
                s sep=2pt,
                inner xsep=4pt,
                inner ysep=5pt,
                line width=1pt,
                ver/.style={rotate=90, child anchor=north, parent anchor=south, anchor=center},
                methodology/.style={fill=oceanblue},
                models/.style={fill=forestgreen},
                evaluation/.style={fill=sunsetorange},
                dataset/.style={fill=rosepink},
                training/.style={fill=goldenyellow},
                efficiency/.style={fill=skyblue},
                application/.style={fill=lavenderblue},
            },
            where level=1{text width=12em,font=\small,}{},
            where level=2{text width=10em,font=\small,}{},
            where level=3{text width=9.5em,font=\small,}{},
            where level=4{text width=25em,font=\scriptsize,}{},
            [
                Instruction-Tuned LLMs, ver 
                [
                    ~Dataset Construction, dataset
                    [
                        ~Human-Annotated, dataset
                        [
                            ~FLAN{,} Natural Instructions{,} SuperNI{,} xP3{,} T0
                            , dataset, text width=24em 
                        ]
                    ]
                    [
                        ~LLM-Generated, dataset
                        [
                            ~Self-Instruct{,} Alpaca{,} WizardLM{,} Evol-Instruct
                            , dataset, text width=24em 
                        ]
                    ]
                    [
                        ~Quality Enhancement, dataset
                        [
                            ~Filtering{,} Deduplication{,} Safety Check{,} Scoring
                            , dataset, text width=24em 
                        ]
                    ]
                ]
                [
                    ~Training Methods, methodology
                    [
                        ~Supervised Fine-Tuning, methodology
                        [
                            ~Standard SFT{,} Multi-Task Learning{,} Curriculum Learning
                            , methodology, text width=24em 
                        ]
                    ]
                    [
                        ~Reinforcement Learning, methodology
                        [
                            ~PPO{,} DPO{,} RRHF{,} Constitutional AI{,} Reward Models
                            , methodology, text width=24em 
                        ]
                    ]
                    [
                        ~Advanced Strategies, methodology
                        [
                            ~Multi-Stage Training{,} Continual Learning{,} Meta-Learning
                            , methodology, text width=24em 
                        ]
                    ]
                ]
                [
                    ~Efficiency Techniques, efficiency
                    [
                        ~Parameter-Efficient, efficiency
                        [
                            ~LoRA{,} AdaLoRA{,} QLoRA{,} P-Tuning{,} BitFit
                            , efficiency, text width=24em 
                        ]
                    ]
                    [
                        ~Data-Efficient, efficiency
                        [
                            ~LIMA{,} AlpaGasus{,} Active Learning{,} Few-Shot
                            , efficiency, text width=24em 
                        ]
                    ]
                    [
                        ~Computational, efficiency
                        [
                            ~DeepSpeed{,} FSDP{,} Mixed Precision{,} Gradient Checkpointing
                            , efficiency, text width=24em 
                        ]
                    ]
                ]
                [
                    ~Model Architectures, models
                    [
                        ~Base Model Families, models
                        [
                            ~GPT{:} InstructGPT{,} ChatGPT{,} GPT-4
                            , models, text width=24em 
                        ]
                        [
                            ~LLaMA{:} Alpaca{,} Vicuna{,} WizardLM{,} LLaMA-2-Chat
                            , models, text width=24em 
                        ]
                        [
                            ~Others{:} FLAN-T5{,} FLAN-PaLM{,} Claude{,} Bard
                            , models, text width=24em 
                        ]
                    ]
                    [
                        ~Specialized Models, models
                        [
                            ~Code{:} StarCoder{,} WizardCoder{,} CodeT5-Instruct
                            , models, text width=24em 
                        ]
                        [
                            ~Math{:} WizardMath{,} MetaMath{,} MAmmoTH
                            , models, text width=24em 
                        ]
                        [
                            ~Multimodal{:} LLaVA{,} InstructBLIP{,} MiniGPT-4
                            , models, text width=24em 
                        ]
                    ]
                ]
                [
                    ~Evaluation Methods, evaluation
                    [
                        ~Automatic Evaluation, evaluation
                        [
                            ~Benchmarks{:} MMLU{,} HellaSwag{,} ARC{,} GSM8K{,} BIG-Bench
                            , evaluation, text width=24em 
                        ]
                    ]
                    [
                        ~Human Evaluation, evaluation
                        [
                            ~Preference{:} Pairwise Comparison{,} Chatbot Arena{,} Likert Scales
                            , evaluation, text width=24em 
                        ]
                    ]
                    [
                        ~LLM-as-Judge, evaluation
                        [
                            ~GPT-4 Eval{,} Claude Eval{,} Multi-Dimensional Scoring
                            , evaluation, text width=24em 
                        ]
                    ]
                ]
                [
                    ~Applications, application
                    [
                        ~Conversational AI, application
                        [
                            ~ChatGPT{,} Claude{,} Bard{,} Character.AI{,} GitHub Copilot
                            , application, text width=24em 
                        ]
                    ]
                    [
                        ~Professional Tools, application
                        [
                            ~Code{,} Writing{,} Translation{,} Analysis{,} Research
                            , application, text width=24em 
                        ]
                    ]
                    [
                        ~Education, application
                        [
                            ~Tutoring{,} Assessment{,} Personalized Learning{,} Q\&A
                            , application, text width=24em 
                        ]
                    ]
                ]
                [
                    ~Challenges \& Future Directions, training
                    [
                        ~Technical Challenges, training
                        [
                            ~Alignment Issues{:} Reward Hacking{,} Goal Misalignment{,} Distributional Shift
                            , training, text width=30em 
                        ]
                        [
                            ~Scalability Problems{:} Training Cost{,} Data Requirements{,} Compute Resources
                            , training, text width=30em 
                        ]
                    ]
                    [
                        ~Safety \& Ethics, training
                        [
                            ~Harmful Content{:} Toxicity Detection{,} Bias Mitigation{,} Content Filtering
                            , training, text width=30em 
                        ]
                        [
                            ~Privacy Concerns{:} Data Protection{,} Model Inversion{,} Federated Learning
                            , training, text width=30em 
                        ]
                    ]
                    [
                        ~Future Research Directions, training
                        [
                            ~Advanced Alignment{:} Constitutional AI{,} Iterated Amplification{,} Debate
                            , training, text width=30em 
                        ]
                        [
                            ~Automated Methods{:} Auto-Instruction{,} Self-Improving Systems{,} Meta-Learning
                            , training, text width=30em 
                        ]
                    ]
                ]
            ]
        \end{forest}
    }
    \caption{A comprehensive taxonomy of Instruction-Tuned Large Language Models, organizing the field into seven key categories: dataset construction, training methods, efficiency techniques, model architectures, evaluation methods, applications, and challenges \& future directions.}
    \label{fig:instruction_tuned_llm_taxonomy}
    \vspace{-2mm}
\end{figure*}
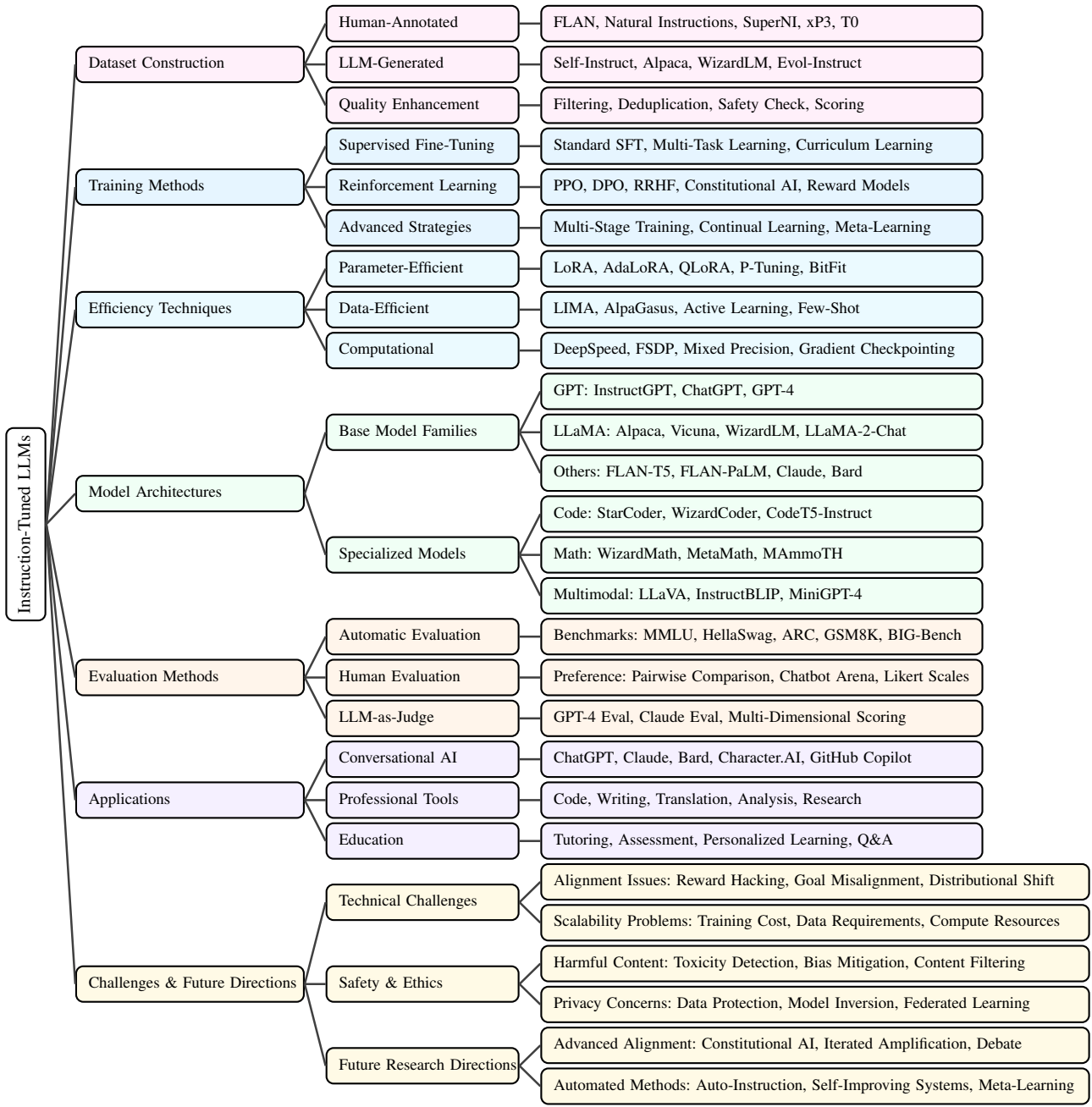

%% file: comp_table1.tex
\FloatBarrier
\begin{table*}[!t]
\centering
\caption{Comprehensive Survey of Instruction-Tuned Large Language Models (Part 1)}
\begin{tabular}{|p{2cm}|p{2cm}|p{3.5cm}|p{3.5cm}|p{3.2cm}|p{0.75cm}|}
\hline
\textbf{Model/Approach} & \textbf{Base Architecture} & \textbf{Fine-tuning Methodology} & \textbf{Key Innovations} & \textbf{Performance Highlights} & \textbf{Year} \\
\hline
\multicolumn{6}{|c|}{\textbf{Commercial Models}} \\
\hline
InstructGPT & GPT-3 (176B) & 1) SFT on human instructions\newline 2) Reward model training\newline 3) PPO optimization & Three-step fine-tuning process using human feedback from API history & +10\% on TruthfulQA, +7\% on toxicity reduction & 2022 \\
\hline
GPT-4 & Next-gen GPT-4 & Large-scale instruction tuning + RLHF & Multimodal with vision and text; improved reasoning and alignment; successor to GPT-3.5 & Top-10\% on bar exam, superior multilingual and safety capabilities & 2023 \\
\hline
Claude 2 & Transformer (100B) & Supervised instruction tuning + Constitutional AI & Introduced constitutional principles to reduce reliance on human feedback & Balanced safety and utility; 100K context window & 2023 \\
\hline
Bard (PaLM 2) & Transformer (PaLM 2) & Supervised fine-tuning + safety alignment & Emphasizes up-to-date knowledge, logic reasoning, and multilingual ability & Strong on open-domain QA, creative writing, and code & 2023 \\
\hline
ERNIE Bot & ERNIE 3.0 (Baidu) & Instruction tuning + knowledge retrieval + human feedback & Integrates knowledge graphs and retrieval; Chinese-optimized & Strong in Chinese QA and multi-turn dialogue & 2023 \\
\hline
Tongyi Qianwen & Qwen Transformer & Instruction tuning + RLHF & Flash Attention, SwiGLU, trained on large Alibaba data & Excels in Chinese/English generation and complex prompts & 2023 \\
\hline
Spark (iFlytek) & Transformer & Multi-stage fine-tuning + RL optimization & Fast/slow thinking training, optimized for domestic hardware & Reasoning + QA + Chinese benchmarks rival GPT-4 & 2023 \\
\hline
Hunyuan & Transformer & Instruction tuning + RLHF + vision tasks & Multimodal, strong long-context and Chinese reasoning & Surpasses GPT-3.5 in many Chinese tasks & 2023 \\
\hline
\multicolumn{6}{|c|}{\textbf{Open-Source Models}} \\
\hline
LIMA & LLaMA & Data-efficient SFT & Strong results from minimal, curated data & Efficient instruction-following with small corpus & 2023 \\
\hline
GPT-4-LLM & LLaMA & N/A & Open-source GPT-4 alternative based on LLaMA & N/A & 2023 \\
\hline
Alpaca & LLaMA 7B & Supervised fine-tuning on synthetic data & Low-cost instruction-following model trained on OpenAI outputs & Achieves usable quality on everyday prompts & 2023 \\
\hline
Vicuna & LLaMA 13B & Fine-tuned on ShareGPT conversations & Distilled from ChatGPT using real multi-turn data & ~90\% ChatGPT quality (GPT-4 judged) & 2023 \\
\hline
Baichuan-13B & Baichuan Transformer & Instruction fine-tuning on bilingual data & Fully open-source, strong Chinese and English performance & Rivals GPT-3.5 on many open benchmarks & 2023 \\
\hline
OpenAssistant & LLaMA 30B & Supervised + RLHF on crowdsourced data & Community-sourced RLHF for open LLMs & ChatGPT-level safety and helpfulness in open source & 2023 \\
\hline
\end{tabular}
\end{table*}

\FloatBarrier 

%% file: comp_table2.tex
\FloatBarrier

\begin{table*}[!t]
\centering
\caption{Comprehensive Survey of Instruction-Tuned Large Language Models (Part 2)}
\begin{tabular}{|p{2cm}|p{2cm}|p{3.5cm}|p{3.5cm}|p{3.2cm}|p{0.75cm}|}
\hline
\textbf{Model/Approach} & \textbf{Base Architecture} & \textbf{Fine-tuning Methodology} & \textbf{Key Innovations} & \textbf{Performance Highlights} & \textbf{Year} \\
\hline
\multicolumn{6}{|c|}{\textbf{Open-Source Models (cont’d)}} \\
\hline
ChatGLM-6B & GLM (6.2B) & Bilingual instruction tuning + quantization & Efficient Chinese-English instruction model runnable on consumer GPUs & Top-tier Chinese model under 10B scale & 2023 \\
\hline
\multicolumn{6}{|c|}{\textbf{Efficient Fine-tuning Methods}} \\
\hline
LoRA & Transformer & Inject trainable low-rank matrices while freezing main weights & Reduces trainable parameters to <0.1\%; modular tuning & Matches full fine-tuning with 0.3\% extra parameters & 2021 \\
\hline
QLoRA & Transformer (4-bit) & 4-bit quantization + LoRA & Enables 65B model tuning on single 48GB GPU & Matches full-tuning on LLaMA-65B; leaderboard topper & 2023 \\
\hline
Prefix Tuning & Transformer & Add trainable KV prefix vectors & Prompt-like control via few learned vectors per layer & Near full-tuning results in few-shot settings & 2021 \\
\hline
Prompt Tuning & Transformer & Train soft prompt embeddings only & No weight updates; all info via learned prompts & 1–2\% gap to full tuning on T5-XXL & 2021 \\
\hline
P-Tuning v2 & Transformer & Deep prompt tuning with layerwise insertion & Better than vanilla prompt tuning for small models & Matches full tuning on SuperGLUE with GPT-3 175B & 2022 \\
\hline
Adapter Tuning & Transformer & Add small bottleneck modules per layer & Isolates tuning to adapters; base remains frozen & 99\% performance using <4\% of parameters & 2019 \\
\hline
BitFit & Transformer & Tune only bias or last layer & Ultra-slim tuning strategy & Comparable classification results, limited for gen tasks & 2021 \\
\hline
IA$^3$ & Transformer & Learn scaling factors per attention/FFN channel & No new matrices; very efficient & Matches LoRA, lower parameter count & 2022 \\
\hline
LLaMA-Excitor & LLaMA & Indirect feature interaction & Preserves pretraining while enhancing instruction compliance & +6\% MMLU improvement & 2024 \\
\hline
Mosaic-IT & N/A & Compositional data augmentation & Diverse instructions without human labeling & 80\% training cost reduction & 2024 \\
\hline
\multicolumn{6}{|c|}{\textbf{Multi-Document Approaches}} \\
\hline
MDCure & N/A & Synthetic + multi-reward tuning & Scalable pipeline without pretraining or human labels & Up to 75.5\% gain vs. baseline & 2024 \\
\hline
DoG-Instruct & N/A & Text-grounded instruction wrapping & Creates semantically grounded instruction data & Boosts multi-doc instruction performance & 2023 \\
\hline
IDEA-MCTS & N/A & Monte Carlo Tree Search & Tree-structured instruction generation framework & General-purpose and scalable approach & 2024 \\
\hline
\multicolumn{6}{|c|}{\textbf{Domain-Specific Models}} \\
\hline
Alpaca & LLaMA & Specialized fine-tuning & Lightweight instruction model for specific domains & Improved performance on narrow domains & 2023 \\
\hline
Vicuna & LLaMA & Specialized fine-tuning & Trained on ShareGPT for nuanced chat tasks & Excellent at multi-turn and domain tasks & 2023 \\
\hline
\multicolumn{6}{|c|}{\textbf{Advanced Approaches}} \\
\hline
Symbolic Task Planner & N/A & Neuro-symbolic integration & Parses instructions into symbolic tasks via neural parsing & Improves generalization and planning & 2024 \\
\hline
\multicolumn{6}{|c|}{\textbf{Dataset Construction Methods}} \\
\hline
Flan & N/A & Dataset integration from multiple NLP tasks & Unifies multi-task data for instruction fine-tuning & Boosts zero/few-shot generalization & 2021 \\
\hline
P3 & N/A & Dataset integration & Similar to Flan; diverse NLP instructions & Broad generalization & 2021 \\
\hline
InstructWild & N/A & LLM-generated synthetic instructions & Diverse synthetic data to expand coverage & Improved instruction coverage & 2023 \\
\hline
Self-Instruct & N/A & LLM-generated & Pipeline for bootstrapping instruction data from LLMs & Expands instruction-tuning corpora & 2023 \\
\hline
\multicolumn{6}{|c|}{\textbf{Evaluation Frameworks}} \\
\hline
LLMEval & N/A & N/A & Benchmarking suite for instruction-following models & Useful for real-world evaluation & 2023 \\
\hline
M-IFEval & N/A & N/A & Evaluation of multilingual instruction adherence & Targets cross-lingual performance & 2025 \\
\hline
\end{tabular}
\end{table*}

\FloatBarrier

%% file: prompt_engineering_table.tex


\newcolumntype{P}[1]{>{\raggedright\arraybackslash}p{#1}}

\onecolumn
\setlength{\tabcolsep}{4pt}   
\renewcommand{\arraystretch}{1.2} 


\begin{longtable}{|P{2.0cm}|P{2.2cm}|P{2.0cm}|P{4.0cm}|P{2.3cm}|P{2.0cm}|}
\caption{\mbox{Prompt Engineering Techniques for Instruction-Tuned LLMs}}
\label{tab:prompt_engineering}\\
\hline
\textbf{Technique} & \textbf{Description} & \textbf{Basic Prompt Example} & \textbf{Enhanced Prompt Example} & \textbf{Key Benefits} & \textbf{Connection to Paper} \\
\hline
\endfirsthead

\multicolumn{6}{c}{{\bfseries Table \thetable\ continued from previous page}} \\
\hline
\textbf{Technique} & \textbf{Description} & \textbf{Basic Prompt Example} & \textbf{Enhanced Prompt Example} & \textbf{Key Benefits} & \textbf{Connection to Paper} \\
\hline
\endhead

\hline \multicolumn{6}{|r|}{{Continued on next page}} \\ \hline
\endfoot

\hline
\endlastfoot

\textbf{Clear Instruction Formatting} & 
Explicit structuring of requests with specific parameters & 
“The advantages and disadvantages of renewable energy.” & 
“Write a balanced analysis of the advantages and disadvantages of renewable energy. Include at least three points for each side and provide concrete examples where possible.” & 
Improves response relevance and completeness; reduces ambiguity & 
Section I – Addresses instruction alignment \\
\hline

\textbf{Role-Based Prompting} & 
Assigning a specific persona to the model & 
“Explain quantum computing to a beginner.” & 
“You are a science educator specializing in explaining complex concepts to beginners. Explain quantum computing to someone with no background in physics or computer science, using simple analogies and avoiding technical jargon.” & 
Establishes appropriate tone, expertise level, and audience adaptation & 
Section I – Leverages model alignment capabilities \\
\hline

\textbf{Chain-of-Thought Prompting} & 
Triggering step-by-step reasoning & 
“A toy costs \$24 with a 20\% discount applied. What was the original price?” & 
“A toy costs \$24 with a 20\% discount applied. Let's solve this step-by-step to find the original price.” & 
Improves accuracy on mathematical and logical problems; enhances transparency & 
Section IV – Relates to reasoning capabilities \\
\hline

\textbf{Few-Shot Learning} & 
Providing examples to establish patterns & 
“Classify: 'Dear Customer, Congratulations! You've been selected to receive a free iPhone. Click here to claim your prize now!'” & 
“Classify the following emails:
Email: 'Meeting tomorrow at 10am.'  
Classification: Not Spam  
Email: 'URGENT: Account compromised!'  
Classification: Spam  
Email: 'Dear Customer, Congratulations! You've been selected...'  
Classification:” & 
Demonstrates expected output format; establishes decision criteria & 
Section II – Connects to data-efficient fine-tuning \\
\hline

\textbf{Multi-Document Analysis} & 
Structured prompting for cross-document tasks & 
“Summarize these two documents about climate change.” & 
“You have two scientific reports about climate change. Analyze both documents and:
1. Identify key findings in both documents  
2. Highlight contradictory conclusions  
3. Synthesize into a cohesive summary  
4. Note methodological differences” & 
Enables complex document comparison; reduces fragmentation & 
Section IV – Implements MDCure-style capabilities \\
\hline

\textbf{Domain-Specific Prompting} & 
Tailoring prompts for specialized knowledge areas & 
“Provide information about treating diabetes in elderly patients.” & 
“You are a healthcare assistant trained to provide evidence-based information. A physician needs information about treatment guidelines for Type 2 Diabetes in elderly patients with kidney complications. Provide:
1. First-line medication recommendations  
2. Monitoring parameters  
3. Dose adjustments for renal impairment  
4. Potential drug interactions” & 
Improves domain relevance; increases precision; enhances utility & 
Section I – Relates to domain-specific applications \\
\hline

\textbf{Hallucination Reduction} & 
Explicit framework for handling uncertain information & 
“Explain the Quantum Gravity Theory developed by physicist Maria Chen.” & 
“If there is a well-documented Quantum Gravity Theory by physicist Maria Chen, explain its key principles. If not, please state this clearly and instead explain the major current approaches to quantum gravity in mainstream physics.” & 
Reduces false information; increases transparency about knowledge limitations & 
Section II – Addresses faithfulness challenges \\
\hline

\textbf{PEFT-Inspired Prompting} & 
Using prompts to emulate specialized fine-tuning & 
“Review this legal contract.” & 
“You are specialized to function as a legal document analyzer with expertise in contract law. When reviewing the contract below, focus exclusively on:
1. Identifying potential liability clauses  
2. Highlighting ambiguous terms  
3. Flagging non-standard provisions” & 
Simulates domain adaptation without technical fine-tuning; focuses model attention & 
Section II – Connects to LoRA and PEFT methods \\
\hline

\textbf{Multilingual Instruction} & 
Prompts designed for cross-lingual evaluation & 
“Explain advantages of electric vehicles.” & 
“Please respond to the following instruction in both English and French, maintaining the same information content and style:
Explain three advantages of electric vehicles compared to traditional combustion engine cars.” & 
Tests multilingual capabilities; ensures cross-lingual consistency & 
Section V – Relates to M-IFEval framework \\
\hline

\textbf{Self-Evaluation Prompting} & 
Instructing the model to assess its own outputs & 
“Solve: A triangle has sides of length 5cm, 12cm, and 13cm. Calculate its area.” & 
“Solve the following mathematical problem step-by-step: A triangle has sides of length 5cm, 12cm, and 13cm. Calculate its area.  
After providing your solution, evaluate your work by:
1. Checking each step  
2. Verifying the final answer  
3. Rating explanation clarity (1–5)  
4. Identifying potential misconceptions” & 
Enhances output quality; provides meta-cognitive analysis; identifies potential errors & 
Section V – Connects to LLM evaluation methods \\
\hline

\textbf{Value-Aligned Prompting} & 
Designing prompts with ethical considerations & 
“Analyze AI for college admissions.” & 
“You are advising a team designing an AI system for college admissions. Provide a balanced analysis of ethical considerations including:
1. Potential sources of bias  
2. Transparency requirements  
3. Accountability mechanisms  
4. Privacy protections  
Consider diverse cultural and socioeconomic perspectives.” & 
Encourages ethical reflection; promotes balanced analysis; reduces harmful outputs & 
Section VII – Relates to ethics and value alignment \\
\hline

\textbf{Compositional Prompting} & 
Combining multiple instructions in structured sequence & 
“Write about climate change solutions.” & 
“First, identify three major contributors to climate change. Then, for each contributor, propose two technological solutions and one policy solution. Finally, analyze which combination of solutions would be most effective, considering implementation challenges and timeframes.” & 
Enables complex multi-step tasks; improves structured responses; enhances logical flow & 
Section IV – Connects to advanced instruction techniques \\
\hline

\textbf{Task Decomposition} & 
Breaking complex problems into manageable subtasks & 
“Create a marketing strategy for a new smartphone.” & 
“To create a marketing strategy for a new smartphone:
1. First, define 3 target demographic segments  
2. For each segment, identify their key priorities when purchasing smartphones  
3. Develop 2 unique selling propositions for each segment  
4. Suggest appropriate marketing channels for each segment  
5. Outline a 3-month launch timeline with key milestones” & 
Improves complex problem solving; enhances thoroughness; makes difficult tasks tractable & 
Section IV – Implements neuro-symbolic approaches \\
\hline

\textbf{Bias Mitigation Prompting} & 
Designing prompts to reduce various biases & 
“Describe a successful entrepreneur.” & 
“Describe five successful entrepreneurs from diverse backgrounds, ensuring representation across different genders, ethnicities, geographic regions, and types of businesses. For each, highlight their unique path to success, challenges they overcame, and distinctive business approaches.” & 
Reduces stereotyping; encourages diverse perspectives; improves representation & 
Section VII – Addresses ethical considerations \\
\hline

\textbf{Contextual Grounding} & 
Providing relevant context before instructions & 
“Summarize the latest research on climate change.” & 
“Context: Different regions experience climate change impacts differently, with varying economic, social and environmental implications.  
Instruction: Summarize the latest IPCC research on climate change, ensuring you address how impacts vary by geographic region and highlighting both global patterns and regional variations.” & 
Increases response relevance; improves accuracy; enhances context-sensitivity & 
Section II – Relates to data curation approaches \\
\hline

\textbf{Output Format Control} & 
Explicitly specifying response structure & 
“Explain the pros and cons of remote work.” & 
“Explain the pros and cons of remote work using the following format:  
\{\#\#\# Advantages\}  
1. [First advantage with brief explanation]  
2. [Second advantage with brief explanation]  
3. [Third advantage with brief explanation]  
\{\#\#\# Disadvantages\}  
1. [First disadvantage with brief explanation]  
2. [Second disadvantage with brief explanation]  
3. [Third disadvantage with brief explanation]  
\{\#\#\# Conclusion\}  
[Balanced summary in 2–3 sentences]” & 
Ensures structured responses; improves readability; facilitates information extraction & 
Section I – Connects to alignment capabilities \\
\hline

\end{longtable}
\twocolumn